\pdfoutput=1

\documentclass[11pt]{article}

\usepackage[final]{acl}

\usepackage{times}
\usepackage{latexsym}

\usepackage[T1]{fontenc}

\usepackage[utf8]{inputenc}

\usepackage{microtype}

\usepackage{inconsolata}

\usepackage{graphicx}

\PassOptionsToPackage{table,xcdraw}{xcolor}
\usepackage{xcolor}   
\usepackage{amsmath}
\usepackage{amssymb}
\usepackage{pifont}
\usepackage{bbm}

\definecolor{ourred}{HTML}{E13342}
\definecolor{ourblue}{HTML}{6495ed}

\newcommand{\rparagraph}[1]{\vspace{1.2mm}\noindent\textbf{#1.}}

\definecolor{Gray}{gray}{0.92}
\definecolor{racing-green}{rgb}{0.0, 0.8, 0.6}
\definecolor{awesome-red}{rgb}{1.0, 0.13, 0.32}

\usepackage{todonotes}
\makeatletter
\newcommand*\iftodonotes{\if@todonotes@disabled\expandafter\@secondoftwo\else\expandafter\@firstoftwo\fi}
\makeatother

\usepackage[inline]{enumitem}

\usepackage{tabularx}
\usepackage{subcaption}
\usepackage{booktabs} 

\usepackage{multirow} 
\usepackage{arydshln}

\usepackage{listings}

\definecolor{stringcolor}{rgb}{0.1,0.5,0.1}
\definecolor{keycolor}{rgb}{0.0,0.0,0.6}
\definecolor{valuecolor}{rgb}{0.6,0.1,0.1}
\definecolor{bggray}{gray}{0.97}
\definecolor{commentgray}{gray}{0.5}

\lstdefinelanguage{json}{
    basicstyle=\ttfamily\small,
    numbers=none,
    backgroundcolor=\color{bggray},
    stepnumber=1,
    showstringspaces=false,
    breaklines=true,
    frame=single,
    rulecolor=\color{gray},
    string=[s]{"}{"},
    stringstyle=\color{stringcolor},
    identifierstyle=\color{keycolor},
    keywordstyle=\color{valuecolor},
    morestring=[b]',
    literate=
     *{:}{{{\color{black}:}}}{1}
      {,}{{{\color{black},}}}{1}
      {[}{{{\color{black}[}}}{1}
      {]}{{{\color{black}]}}}{1}
      {\{}{{{\color{black}\{}}}{1}
      {\}}{{{\color{black}\}}}}{1},
}

\usepackage[most]{tcolorbox}
\tcbuselibrary{listings, skins}

\newtcolorbox[use counter=lstlisting]{examplebox}[2][]{%
  colback=gray!20,
  colframe=black,
  width=\linewidth,
  boxsep=5pt,
  left=2pt,
  right=2pt,
  top=2pt,
  bottom=2pt,
  fonttitle=\bfseries,
  title={Example~\thelstlisting: #2},
  label={#1},
  enhanced,
  breakable,
  }

\definecolor{codebg}{rgb}{0.98,0.98,0.98}
\definecolor{keyword}{RGB}{0,0,255}
\definecolor{comment}{RGB}{0,128,0}
\definecolor{string}{RGB}{163,21,21}
\definecolor{outputcolor}{RGB}{128, 0, 128}

\title{Quantifying Language Disparities in Multilingual Large Language Models}

\author{
  Songbo Hu~~~ 
  Ivan Vuli\'{c}~~~
  Anna Korhonen
  \\
  Language Technology Lab, University of Cambridge, UK
  \\
  \texttt{\{sh2091,iv250,alk23\}@cam.ac.uk} \\
}

\begin{document}
\maketitle
\begin{abstract}
Results reported in large-scale multilingual evaluations are often fragmented and confounded by factors such as target languages, differences in experimental setups, and model choices. We propose a framework that disentangles these confounding variables and introduces three interpretable metrics---the performance realisation ratio, its coefficient of variation, and language potential---enabling a finer-grained and more insightful quantification of actual performance disparities across both (i) models and (ii) languages. Through a case study of 13 model variants on 11 multilingual datasets, we demonstrate that our framework provides a more reliable measurement of model performance and language disparities, particularly for low-resource languages, which have so far proven challenging to evaluate. Importantly, our results reveal that higher overall model performance does not necessarily imply greater fairness across languages.
\end{abstract}

\section{Introduction}
Contemporary NLP development relies on digital language datasets to build large language models (LLMs) and adapt them to downstream applications. Yet, none of these resources offers a balanced representation of the world’s more than 7,000 languages~\cite{blasi-etal-2022-systematic, joshi-etal-2020-state, 10.1145/3442188.3445922}. This imbalance introduces two types of bias: \textit{intrinsic bias}, which shapes the LLM’s internal representations; and \textit{adaptation bias}, which affects downstream performance after task-specific adaptation~\cite{bommasani2021opportunities}. Given the common assumption of a strong correlation between the quantity of in-language data and LLM performance on that language, adaptation bias is often considered \textit{observable}, as the amount of task-specific data is typically known; see, e.g.,~\cite{hu-etal-2023-systematic}. In contrast, intrinsic bias in LLMs is more latent, becoming apparent only during evaluation on downstream tasks under parallel settings.

Under the current paradigm, the evaluation of multilingual LLMs is often \textit{fragmented}, in the sense that reported scores are typically obtained on isolated datasets, each covering different sets of languages and often targeting different NLP tasks. This leads to several limitations: (i) No existing dataset covers more than a small fraction of human languages, leading to severe and persistent under-representation. (ii) Reported scores are often confounded by multiple variables, such as the choice of languages, training and evaluation setups, making it difficult to compare results across languages, datasets, and metrics. (iii) Consequently, most findings from such multilingual evaluations remain qualitative, typically concluding that performance is uneven across languages, see e.g.~\cite{pmlr-v119-hu20b, xuan2025mmlu}, without quantifying the magnitude of these disparities. (iv) Evaluating models on isolated datasets can lead to under- or over-estimation of performance for low-resource languages (see later in \S\ref{sec:case_study}), thereby hindering efforts to promote fairness across languages in NLP.

To mitigate the aforementioned limitations, we propose a framework that disentangles confounding factors in multilingual evaluations, enabling the quantification of language disparities using evidence from multiple datasets.  Drawing inspiration from measurement theory (\S\ref{sec:rw}), we define a latent construct called \textit{performance potential}, which estimates the score a model is expected to achieve on a given task for a specific language, given the current state of NLP. Here, we adopt a norm-referenced approach (i.e., using the aggregate scores of the sampled population of LLMs as a reference). Then, we introduce the \textit{performance realisation ratio}, defined as the ratio between the actual score achieved by the model and the estimated potential. Finally, we quantify model-level fairness using the \textit{coefficient of variation} of this ratio across languages, which captures the relative dispersion in performance realisation across languages. Through a case study, we demonstrate that our proposed framework provides a more reliable measurement, particularly for low-resource languages. We showcase the core analysis code in Example~\ref{list:code} in the Appendix, and release the full code and data at: \href{https://github.com/cambridgeltl/quantifying_language_disparities}{\nolinkurl{github.com/cambridgeltl/quantifying_language_disparities}}.

\section{Related Work}
\label{sec:rw}

\noindent \textbf{Measuring Performance as a Construct.}  
Evaluation of language models can be viewed as a process of assigning numerical values to models, and thus naturally falls within the scope of \textit{measurement theory} (see introductions in~\citealp{hand2016measurement, bandalos2018measurement}). Measurement theory emphasises the distinction between a \textit{construct}, the abstract concept we aim to measure, and the specific procedure used to produce its numerical representation. In our case, performance is the construct, while the reported scores serve as its representation. Tools from measurement theory have been increasingly adopted in NLP, including efforts to assess reliability~\cite{zhou2024larger} and robustness~\cite{hardy-2025-glitters} of LLMs, examine both the reliability and validity of NLG evaluation metrics~\cite{xiao-etal-2023-evaluating-evaluation}, reduce evaluation costs through adaptive testing~\cite{sedoc-ungar-2020-item}, and analyse dataset difficulty~\cite{bachmann2024fl, tambon2024taskeval}. Our framework draws inspiration from variance decomposition techniques used in generalisability theory~\cite{cronbach1972dependability}, which estimate the contributions of various measurement facets to observed score variability. We transpose and adapt this idea to multilingual evaluation.

\vspace{0.5mm}
\noindent \textbf{Definition of Fairness.}
Comparing the performance of multilingual LLMs can lead to a dilemma, where no single model consistently outperforms others across all languages. In such cases, aggregating multilingual performance into a single number for model selection inevitably challenges prevailing notions of fairness~\cite{Choudhury_Deshpande_2021}. Similarly, cross-lingual comparisons also implicitly reflect on particular assumptions about fairness. Many multilingual NLP efforts aim to equalise performance across languages. This reflects a principle aligned with \textit{equal outcome}. However, given the current disparities in digital resources across languages, such a goal is often difficult to operationalise. In this work, we adopt the \textit{broader view of equal opportunity}, as discussed by~\citet{barocas2023fairness}. Specifically, we define a fair LLM as one that is constructed in a way that enables all languages to realise their \textit{performance potential}, that is, to achieve the level of performance that is realistically achievable given the current state of available resources.

\section{Measurement Framework}
\label{sec:method}
We propose a two-step framework: (i) estimating the \textit{performance potential} for each language-task pair; and (ii) quantifying language disparities at both the language and model levels.

\rparagraph{Preliminaries}
The input to our framework is a set of evaluation records, each represented as a tuple $(\ell, t, m, s) \in \mathcal{D}$, where $\ell$ denotes a language, $t$ denotes an evaluation task defined as a dataset–metric pair, $m$ denotes a language model, and $s \in \mathbb{R}$ is the observed performance score. Here, $\ell$, $t$, and $m$ are treated as categorical variables. 

\rparagraph{Step 1: Estimating Performance Potential}
To isolate the achievable performance of a language from task difficulty and model capabilities, we model the observed score using a linear mixed-effects model as follows, which aligns with those widely used in generalisability theory~\cite{BrennanRobertL2001GT}:
\begin{equation}
s_{\ell,t,m} = \mu + \alpha_{\ell} + \beta_t + u_m + \varepsilon_{\ell,t,m}
\label{eq:lmm}
\end{equation}
\noindent
where $\mu$ is the intercept; $\alpha_{\ell}$ and $\beta_t$ are the language and task-specific fixed effects, respectively; $u_m \sim \mathcal{N}(0, \sigma^2_u)$ is a random effect for model $m$; and $\varepsilon_{\ell,t,m} \sim \mathcal{N}(0, \sigma^2)$ is the residual noise term. We then define the \textit{performance potential} for a given language-task pair as the marginal mean score, averaged over the model population:
\begin{equation}
\mathrm{PP}_{\ell,t} := \widehat{\mathbb{E}}_m \left[ s_{\ell,t,m} \right] = \mu + \alpha_{\ell} + \beta_{t}
\label{eq:pp}
\end{equation}
\noindent
where the expectation is taken over the random effect $u_m$ with $\mathbb{E}[u_m]=0$.

\rparagraph{Step 2: Quantifying Language Disparities}
At the level of languages, we define the \textit{language potential} as the performance of a typical model on a typical task, given current NLP resources:
\begin{equation}
\mathrm{LP}_\ell := \widehat{\mathbb{E}}_{t} \left[ \mathrm{PP}_{\ell,t} \right] = \mu + \alpha_{\ell} + \operatorname{Mean}_{t} (\beta_t)
\label{eq:lp}
\end{equation}
where $\operatorname{Mean}_{t} (\beta_t)$ denotes the mean task effect across all observed tasks. This quantity provides an aggregate indication of how well a language is currently served by NLP models. At the level of models, we define the \textit{performance realisation ratio} for each $(\ell, t, m)$ as:
\begin{equation}
\mathrm{PRR}_{\ell,t,m} := \frac{s_{\ell,t,m}}{\mathrm{PP}_{\ell,t}}
\label{eq:prr}
\end{equation}
\noindent This ratio can be interpreted as the normalised performance of model $m$ on language $\ell$ and task $t$, relative to the estimated achievable score for that language-task pair. Finally, we measure \textit{model-level language disparities} using the coefficient of variation of PRR across languages (and tasks):
\begin{equation}
\mathrm{CV}^{(m)}_{\mathrm{PRR}} := 
\frac{
\operatorname{Std}_{(\ell, t)} \left( \mathrm{PRR}_{\ell, t, m} \right )
}{
\operatorname{Mean}_{(\ell, t)} \left( \mathrm{PRR}_{\ell, t, m} \right )
}
\label{eq:cvprr}
\end{equation}
where $\operatorname{Std}_{(\ell, t)}$ and $\operatorname{Mean}_{(\ell, t)}$ denote the standard deviation and mean computed over all available language-task pairs $(\ell, t)$ for a fixed model $m$. The denominator, $\operatorname{Mean}_{(\ell, t)} \left( \mathrm{PRR}_{\ell, t, m} \right)$, can also serve as an evaluation metric, indicating the model's overall normalised performance.

\section{Case Study}
\label{sec:case_study}

\rparagraph{Experimental Setup}
We demonstrate the potential of our framework using published results from the MEGA benchmark~\cite{ahuja-etal-2023-mega}.\footnote{Our method is applicable both to primary experimental results (e.g., evaluating LLMs on benchmarks) and to meta-analyses, where published results from other studies are used as secondary data. In this case study, we focus on the latter: meta-evaluating previously published results.} The analysis covers 13 multilingual models and model variants from two categories: (i) LLMs with in-context learning and (ii) fine-tuned models. For LLMs, we analyse results under the default monolingual prompting setting (where both the test input and few-shot exemplars are in the target language), and, for OpenAI models, additionally under the translate-test (TT) setup (where test examples are translated into English and English exemplars are used). We evaluate these models on 11 multilingual datasets included in the MEGA benchmark. Details are provided in Appendix~\ref{sec:setup}, and the data used are listed in Appendix~\ref{sec:mega_result_appendix}. For each reported score in the benchmark, we extract the evaluation records for each model, language, dataset, metric used, and corresponding evaluation score. Each dataset-metric pair is treated as a single evaluation task within our framework. In total, we obtained 1,364 evaluation records.

\begin{figure}[!t]
    \centering
    \includegraphics[width=0.8\linewidth]{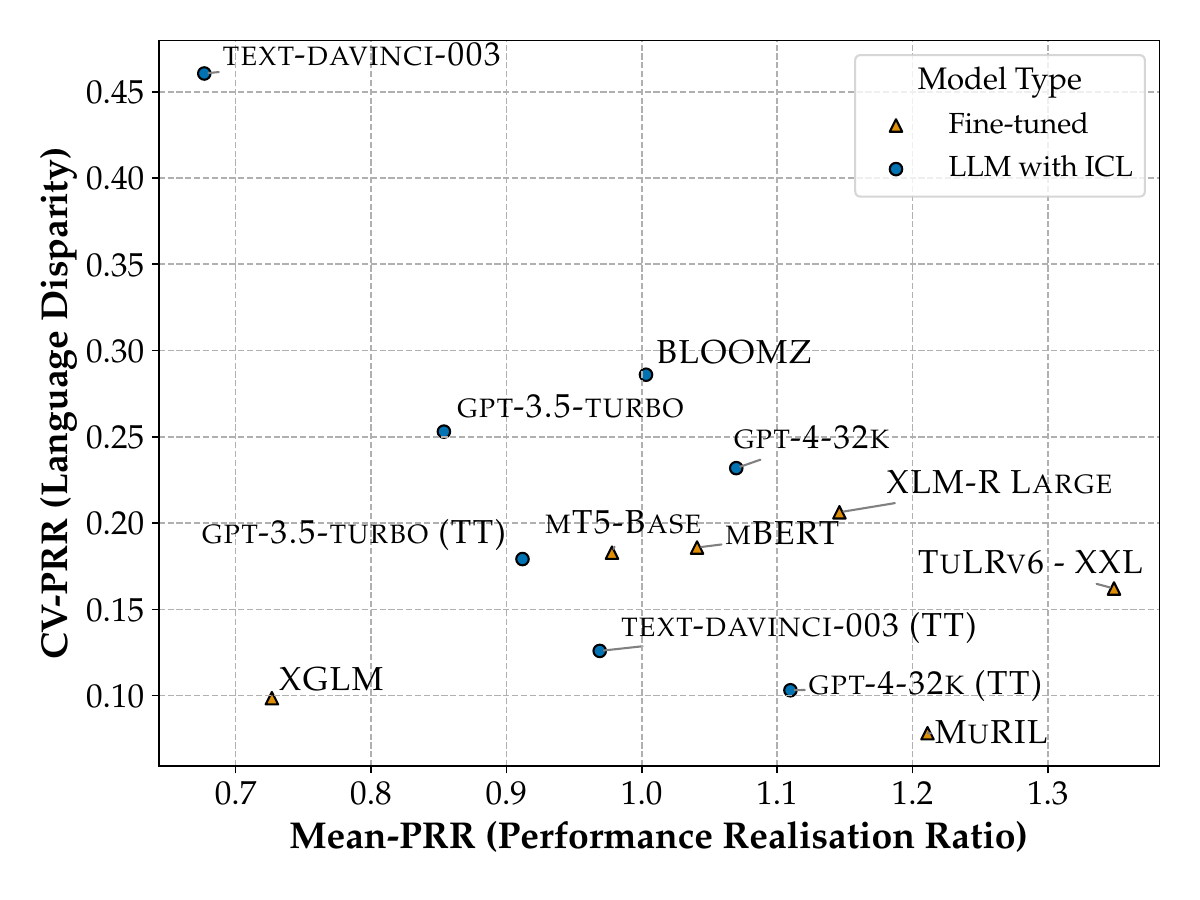}
    \caption{
        Model performance, measured by the mean performance realisation ratio (Mean-PRR), and model-level language disparities, measured by the coefficient of variation of PRR (CV-PRR). Higher Mean-PRR indicates better overall performance. Lower CV-PRR reflects reduced disparities. Results are based on Table~\ref{tab:model_results_appendix}.
    }
    \label{fig:main_result}
\end{figure}

\rparagraph{Results}
Figure~\ref{fig:main_result} presents the results of the model-level evaluation based on our framework. We observe that models with higher overall performance do not necessarily show less disparity across languages. For example, fine-tuned \texttt{mT5-Base} and \texttt{mBERT} models achieve lower Mean-PRR compared to \texttt{gpt-4}, but also achieve lower CV-PRR, indicating lower language disparities across languages. Additionally, when comparing the OpenAI model family (i.e., \texttt{text-davinci-003}, \texttt{gpt-3.5-turbo}, and \texttt{gpt-4}), we observe a trend in which larger, more recent, and arguably more advanced models achieve both better overall performance and reduced language disparities, indicating some progress in mitigating existing language inequities.

\begin{figure}[t]
    \centering
    
    \begin{subfigure}{0.9\linewidth}
        \centering
        \includegraphics[width=\linewidth]{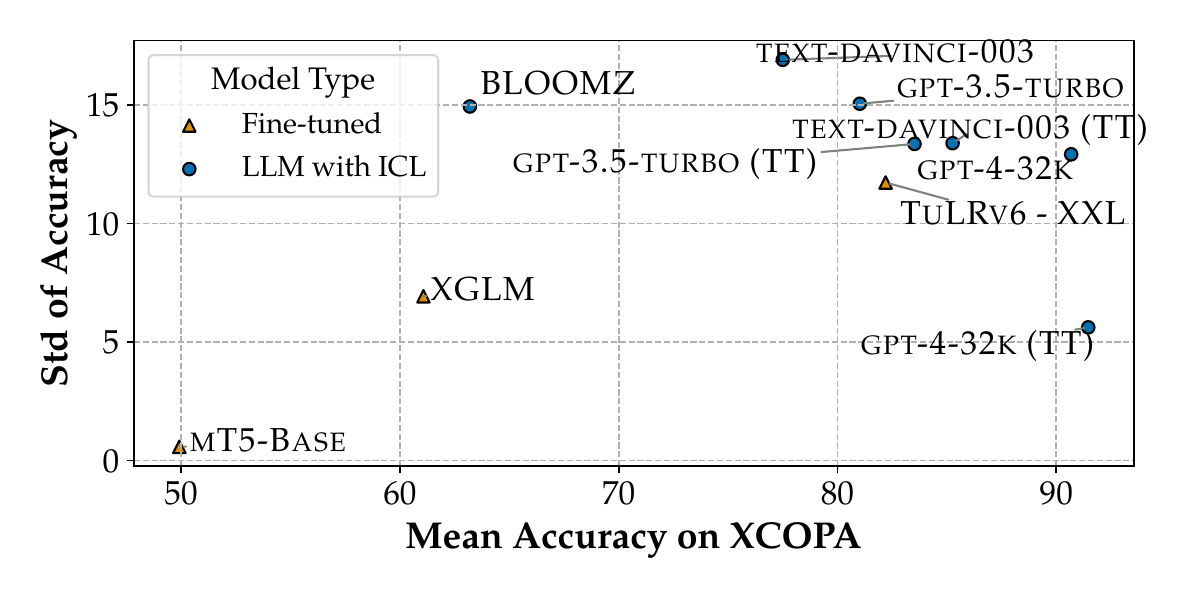}
    \end{subfigure}

    \vspace{-2em}
  
    \begin{subfigure}{0.86\linewidth}
        \centering
        \includegraphics[width=\linewidth]{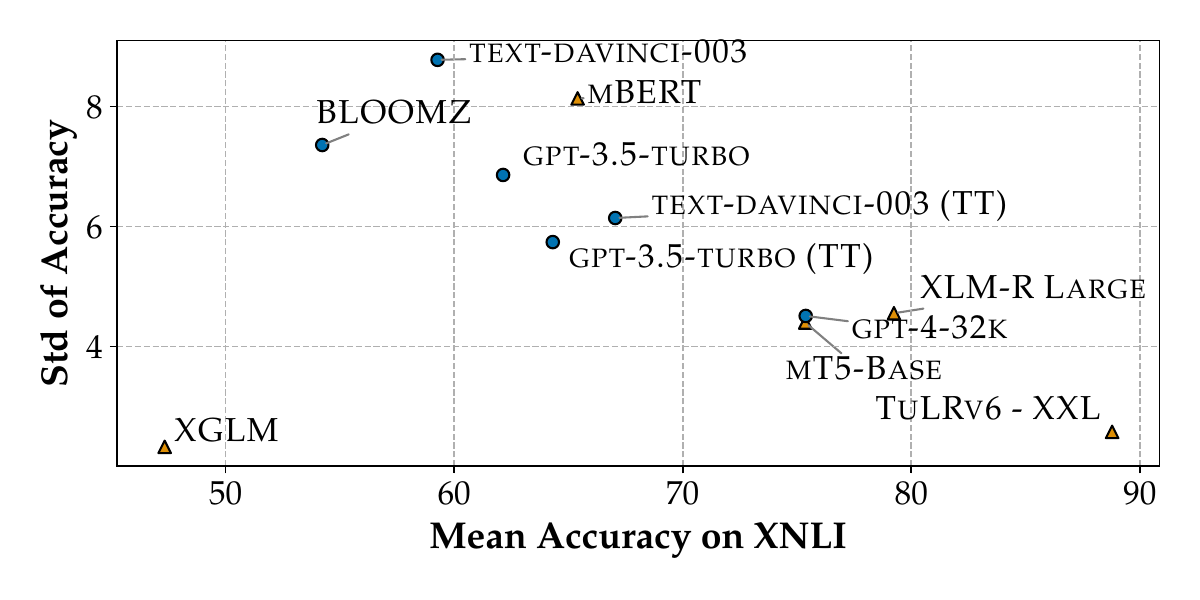}
    \end{subfigure}

    \vspace{-2em}
    
    \begin{subfigure}{0.86 \linewidth}
        \centering
        \includegraphics[width=\linewidth]{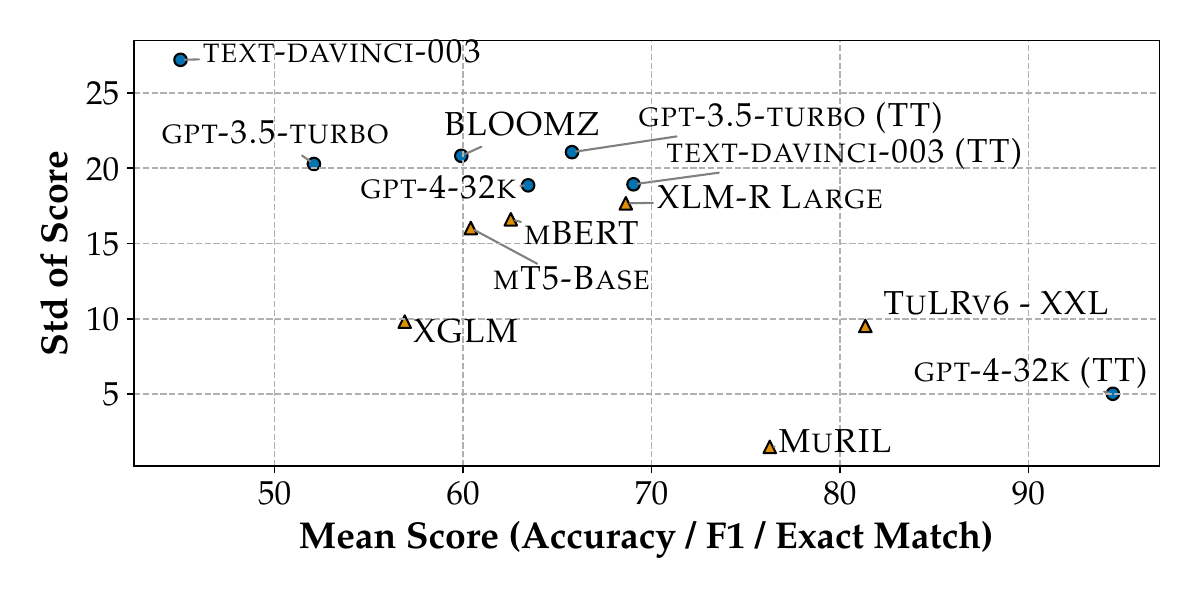}
    \end{subfigure}
    \caption{Baseline evaluation of model performance (based on the mean reported score) and language disparity (based on the standard deviation of reported scores) on (top) the XCOPA dataset, (middle) the XNLI dataset, and (bottom) across all datasets and metrics combined.}
    \label{fig:compare_result_all}
\end{figure}

For comparison, the bottom subfigures of Figure~\ref{fig:compare_result_all} show a baseline aggregation of results across different evaluation setups, calculated by taking the mean and standard deviation without accounting for confounding factors. 
This approach yields evaluations that differ from those produced by our proposed framework. It is evident that such a common baseline can be problematic, as illustrated by the following scenario: suppose we evaluate only the OpenAI models, excluding others, on a particularly challenging dataset. In this case, their average performance would drop significantly. Without reference to the performance of other models, this decrease in the mean score would not provide meaningful insight.

A major benefit of our proposed framework is its ability to disentangle factors that are often conflated in multilingual evaluations, thereby resolving a key challenge in multilingual NLP: enabling meaningful comparisons across datasets and metrics. For example, the top and middle subfigures of Figure~\ref{fig:compare_result_all} show model evaluation results on the XNLI and XCOPA datasets. These results highlight a common dilemma: when comparing model performance across different datasets or metrics, the ranking of models is often inconsistent due to such confounding effects.

\begin{figure}[!t]
    \centering
    \includegraphics[width=0.8\linewidth]{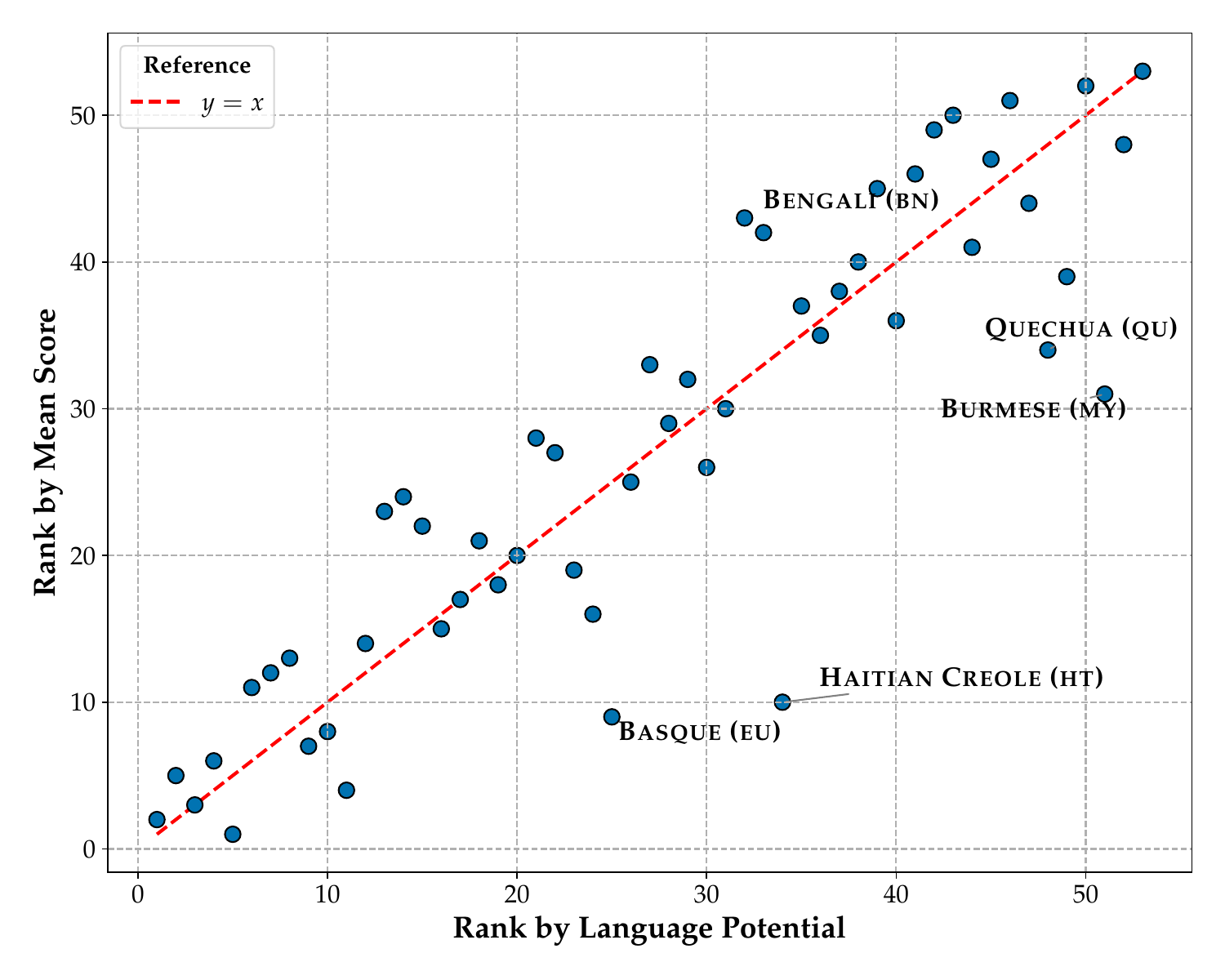}
    \caption{Ranking of languages based on language potential (ours), compared to the average scores reported for all models and datasets for each language (baseline). The top 5 languages with the most divergent rankings (i.e., most misestimated by the baseline) are labelled. Results are based on Table~\ref{tab:full_langpot_meanscore}.}
    \label{fig:langauge_rank}
\end{figure}

Our framework can also be applied to estimate how well a language is served by current models. Figure~\ref{fig:langauge_rank} presents a ranking of languages according to their estimated language potential (as defined in Equation~\ref{eq:lp}), and compares this to a baseline ranking based on the average score reported for all models and datasets for each language. While the two rankings show a roughly linear correlation, the baseline method systematically misestimated the ranking of several low-resource languages. These cases arise from the uneven coverage of languages across datasets, a limitation that disproportionately affects low-resource languages. For example, Haitian Creole (HT) is only covered in XCOPA, a dataset on which models achieve relatively high performance (e.g., \texttt{gpt-4} achieves 99.6 in English). As a result, the baseline approach ranks Haitian Creole in the top 10, whereas our framework places it at 34. By disentangling the confounding effect of task difficulty, our framework thus provides a more representative overview for low-resource languages.

\rparagraph{Diagnosis}
We conducted model diagnostics on the assumptions of the linear mixed-effects model and performed robustness checks of our proposed framework (see details in Appendix~\ref{sec:diagnostics_appendix}). Visual inspection of Figures~\ref{fig:qq_plot_residual_appendix}-\ref{fig:qq_plot_random_appendix} indicates that our data generally satisfy these assumptions, although some deviations from ideal model assumptions, such as the normality and homoscedasticity of residuals, are observed.  We attribute these deviations primarily to the simplicity of the models used (as the linear model does not yet account for interactions between effects). Our framework can readily be extended to more complex statistical models as more data become available. Importantly, our robustness tests show that model and language rankings (see Figures~\ref{fig:mean_prr_robust_appendix} and~\ref{fig:cv_prr_robust_appendix}), remain highly consistent even after the removal of outliers. This evidence supports the reliability of our framework for evaluating LLMs in realistic settings where perfect experimental control is rarely achievable. We therefore present our framework as a practical tool for multilingual evaluation.

\rparagraph{Generalisability Across Benchmarks and Tasks}
Our framework is designed to be dataset-agnostic: any set of model--task--language evaluation scores can be incorporated, and larger, more diverse samples yield more reliable estimates, similar to other statistical approaches. To test its generalisability, we extended our analysis beyond MEGA to two additional multilingual benchmarks: XTREME-R~\cite{ruder-etal-2021-xtreme} (10 datasets, 50 languages) and M5~\cite{schneider-sitaram-2024-m5} (8 multilingual vision–language datasets, 41 languages). The resulting language potential rankings remained highly consistent with those derived from MEGA (Spearman $\rho=0.87$, Kendall $\tau=0.73$ for 50 languages shared between MEGA and XTREME-R; and Spearman $\rho=0.79$, Kendall $\tau=0.62$ for 29 languages shared between MEGA and M5; all $p < 0.001$). Despite differences in tasks, modalities, and evaluated models, these strong correlations confirm that our method provides robust and generalisable cross-lingual insights. Moreover, the framework is not tied to any specific metric or task, and can be applied across diverse performance measures. For instance, Table~\ref{tab:m5_results_appendix} in the Appendix shows the model-level evaluation results based on our framework for the M5 benchmark, which focuses on multilingual and multicultural vision–language tasks. These results illustrate that our framework is not limited to text-only benchmarks, but can be generalised to other evaluation settings as well.

\section{Conclusion}
\label{sec:conclusion}
We propose a measurement framework to disentangle confounding factors in multilingual evaluation, thereby enabling the quantification of language disparities. Our framework introduces interpretable metrics, including the performance realisation ratio (PRR), its coefficient of variation (CV-PRR), and language potential, which facilitates reliable comparisons across models and languages. Through a case study, we demonstrate that our approach yields more equitable and robust evaluations, particularly for low-resource languages. Notably, our findings reveal that higher overall model performance does not guarantee greater cross-lingual fairness. We believe our framework can serve as a practical tool for multilingual evaluation, with the potential for broader application as more data becomes available.

\section*{Limitations}
\label{sec:limitation}
Our proposed framework relies on linear mixed-effects models, which impose several classical statistical assumptions: linearity, homoscedasticity, and normality. Linearity here implies that the effects of language, model, and task are additive, and interaction effects (e.g., language–model interactions) are not explicitly modelled. This is a reasonable and intuitive assumption for multilingual model evaluation, where strong models are expected to perform better across a wide range of languages and tasks. However, the assumptions of homoscedasticity and normality are less straightforward in practical settings. Homoscedasticity requires that the residual variance remains constant across languages and tasks; yet, this assumption can be violated in multilingual NLP when evaluation setups are not fully parallel across languages, particularly for low-resource cases (see Table~\ref{tab:app_udpos} for an example). Similarly, the normality assumptions for both residuals and random effects may be violated in practice, particularly in the presence of outliers.

In our model diagnostics in \S\ref{sec:case_study}, we acknowledge that the assumptions of homoscedasticity and normality of the residuals are statistically rejected, while other model assumptions are satisfied. However, we argue that these violations do not fundamentally undermine the validity or utility of our framework. Even when certain model assumptions are not fully met, all reported results and analyses, such as model rankings, remain entirely valid with respect to the observed data samples. The main implication of these violations is that caution should be exercised when generalising these findings to other datasets or to the broader model population. This is a limitation not unique to our framework, but common to most evaluation metrics for which generalisability is seldom explicitly tested.

Moreover, these issues could be mitigated by employing more sophisticated models, such as those incorporating interaction terms (e.g., task-language interactions), which allow the model to capture the fact that certain tasks may be more challenging for specific languages. 
For instance, Equation~\ref{eq:lmm} can be extended as:
\begin{equation}
s_{\ell,t,m} = \mu + \alpha_\ell + \beta_t + \gamma_{\ell,t} + u_m + \varepsilon_{\ell,t,m}
\label{eq:interaction_term}
\end{equation}
where $\gamma_{\ell,t}$ captures language--task interaction effects.
Our framework supports such extensions, which can in turn enable more rigorous validation of the assumptions underlying the statistical model in future work.
If such assumptions can be statistically validated, evaluation results produced by our framework will be supported not only by intuitive reasoning but also by formal statistical evidence, an important and rarely achieved attribute for evaluation metrics.

The main reason we did not use a more complex model in this paper is due to the limitation of available data: many existing multilingual benchmarks do not provide language-level breakdowns of performance for each model and instead report results averaged across languages. We therefore encourage future multilingual evaluations to include per-language performance breakdowns in their published results, which would enable more fine-grained and robust statistical modelling in future research. This data constraint also leads to another limitation of this paper: our case study is based solely on the reported scores from the MEGA paper. However, we argue that the comprehensive results provided by the MEGA benchmark, which cover 11 datasets and 13 models and model variants, offer a realistic and representative setting to evaluate our framework.

Finally, while we provide evidence for the reliability of our framework as a measurement tool through model diagnostics and robustness tests, these analyses alone do not show its validity. In our context, validity refers to whether the performance and disparities measured by our framework truly reflect the underlying constructs of interest (e.g., whether lower disparities actually correspond to more equitable or fairer utility for end users). Demonstrating such validity typically requires external evidence and is particularly challenging in NLP, where the link between evaluation metrics (e.g., accuracy or F1 score) and real-world utility is often assumed rather than empirically validated. Like most existing approaches, our framework implicitly assumes that higher reported scores correspond to greater utility. While we acknowledge this as a limitation, it remains a pragmatic and widely adopted assumption in current NLP research. Notably, our framework improves upon many existing metrics by providing a principled approach with a clear definition of fairness.

\section*{Acknowledgements}
Songbo Hu is supported by the Cambridge International Scholarship.
Ivan Vuli\'{c} acknowledges the support of a personal Royal Society University Research Fellowship \textit{`Inclusive and Sustainable Language Technology for a Truly Multilingual World'} (no 221137; 2022--). 
Anna Korhonen acknowledges the support of the UKRI Frontier grant EP/Y031350/1.

\bibliography{custom}

\appendix
\section{Experimental Setup for Case Study}
\label{sec:setup}

Our analysis covers two types of multilingual models: (i) LLMs with in-context learning, including OpenAI’s \texttt{text-davinci-003}, \texttt{gpt-3.5-turbo}, \texttt{gpt-4}~\cite{achiam2023gpt}, and \texttt{BLOOMZ}~\cite{muennighoff-etal-2023-crosslingual}; and (ii) smaller fine-tuned multilingual models, such as \texttt{mBERT}~\cite{devlin-etal-2019-bert}, \texttt{XLM-R}~\cite{conneau-etal-2020-unsupervised}, \texttt{mT5}~\cite{xue-etal-2021-mt5}, \texttt{MuRIL}~\cite{khanuja2021muril}, and \texttt{TULRv6}~\cite{patra-etal-2023-beyond}. We analyse these models on broad range of NLP tasks, including natural language inference (XNLI~\cite{conneau-etal-2018-xnli}, IndicXNLI~\cite{aggarwal-etal-2022-indicxnli}), commonsense reasoning (XCOPA~\cite{ponti-etal-2020-xcopa}, XStoryCloze~\cite{lin-etal-2022-shot}), paraphrase identification (PAWS-X~\cite{yang-etal-2019-paws}), question answering (XQuAD~\cite{artetxe-etal-2020-cross}, MLQA~\cite{lewis-etal-2020-mlqa}, TyDiQA-GoldP~\cite{clark-etal-2020-tydi}, IndicQA~\cite{doddapaneni-etal-2023-towards}), and sequence labelling (PAN-X~\cite{pan-etal-2017-cross}, UDPOS~\cite{nivre2018universal}). Each dataset is evaluated across multiple languages. For each dataset, we adopt the evaluation metrics reported in the MEGA benchmark (e.g., accuracy and F1), and treat each dataset–metric pair as a single evaluation task in our framework. For further details about the models, datasets, and prompting strategies, we refer readers to the original MEGA benchmark paper~\cite{ahuja-etal-2023-mega}.

\vspace{0.5mm}
\noindent
\textbf{Example~\ref{list:code}} provides the core code used to conduct the analyses in our framework. The data utilised for our case study is provided in \S\ref{sec:mega_result_appendix} of the Appendix. The code snippet is released under the MIT License.

\begin{lstlisting}[language=Python, basicstyle=\footnotesize\ttfamily, breaklines=true, caption={Core analysis code for our proposed framework.}, label={list:code}]
import pandas as pd
import numpy as np
import statsmodels.formula.api as smf

# Example input (JSON entry in model_performance_record_list.json)
# [
#   {"Model": "gpt-3.5-turbo", "Language": "en", "Score": 97.8, "Dataset": "xcopa", "Metric": "accuracy"},
#   ...
# ]

df = pd.read_json("model_performance_record_list.json")

df["Task"] = df["Dataset"] + "_" + df["Metric"]
for col in ["Language", "Task", "Model"]:
    df[col] = df[col].astype("category")

md = smf.mixedlm("Score ~ C(Language) + C(Task)", df, groups=df["Model"], re_formula="~1")
mdf = md.fit(method="lbfgs", reml=False)

intercept = mdf.params["Intercept"]
language_effect = mdf.params.filter(like="C(Language)").to_dict()
task_effect = mdf.params.filter(like="C(Task)").to_dict()
mean_task_effect = np.mean(list(task_effect.values()))

def alpha_lang(lang): return language_effect.get(f"C(Language)[T.{lang}]", 0.0)
def beta_task(task): return task_effect.get(f"C(Task)[T.{task}]", 0.0)

lang_task_potential = {
    (lang, task): intercept + alpha_lang(lang) + beta_task(task)
    for (lang, task) in df[["Language", "Task"]].drop_duplicates().itertuples(index=False)
}
df["Potential"] = df.apply(lambda r: lang_task_potential[(r.Language, r.Task)], axis=1)

language_potential = {
    lang: intercept + alpha_lang(lang) + mean_task_effect
    for lang in df["Language"].cat.categories
}
language_df = (
    pd.DataFrame(language_potential.items(), columns=["Language", "Potential"])
      .sort_values("Potential", ascending=False)
)

df["PRR"] = df["Score"].astype(float) / df["Potential"]

model_evaluation = df.groupby("Model")["PRR"].agg(mean_prr="mean", std_prr="std").assign(
    cv_prr=lambda x: x["std_prr"] / x["mean_prr"]
).reset_index()

print("Language Potential")
print(language_df.to_string(index=False))

print("Model-Level Evaluation")
print(model_evaluation.to_string(index=False))
\end{lstlisting}

\section{Supplementary Results for Case Study}
\label{sec:additional_result}

\textbf{Table~\ref{tab:model_results_appendix}} shows the model-level evaluation results based on our framework for the case study presented in \S\ref{sec:case_study}, as used in Figure~\ref{fig:main_result}.

\begin{table}[ht]
    \centering
    \resizebox{\columnwidth}{!}{
    \begin{tabular}{lccc}
        \toprule
        Model & Mean-PRR & Std-PRR & CV-PRR \\
        \midrule
        BLOOMZ                & 1.00 & 0.29 & 0.29 \\
        MuRIL                 & 1.21 & 0.09 & 0.08 \\
        TuLRv6 - XXL          & 1.35 & 0.22 & 0.16 \\
        XGLM                  & 0.73 & 0.07 & 0.10 \\
        XLM-R Large           & 1.15 & 0.24 & 0.21 \\
        gpt-3.5-turbo         & 0.85 & 0.22 & 0.25 \\
        gpt-3.5-turbo (TT)    & 0.91 & 0.16 & 0.18 \\
        gpt-4-32k             & 1.07 & 0.25 & 0.23 \\
        gpt-4-32k (TT)        & 1.11 & 0.11 & 0.10 \\
        mBERT                 & 1.04 & 0.19 & 0.19 \\
        mT5-Base              & 0.98 & 0.18 & 0.18 \\
        text-davinci-003      & 0.68 & 0.31 & 0.46 \\
        text-davinci-003 (TT) & 0.97 & 0.12 & 0.13 \\
        \bottomrule
    \end{tabular}
    }
        \caption{
        Model-level evaluation results based on our framework. 
        \textbf{Mean-PRR}: Mean performance realisation ratio; 
        \textbf{Std-PRR}: Standard deviation of the PRR; 
        \textbf{CV-PRR}: Coefficient of variation of the PRR.
    }
    \label{tab:model_results_appendix}
\end{table}

\vspace{0.5mm}
\noindent
\textbf{Table~\ref{tab:full_langpot_meanscore}} shows the language potential calculated in \S\ref{sec:case_study}, as used in Figure~\ref{fig:langauge_rank}.

\begin{table}[ht]
\centering
\resizebox{\columnwidth}{!}{
\begin{tabular}{lcccc}
\toprule
\textbf{Language (Code)} & \textbf{Rank (Potential)} & \textbf{Potential} & \textbf{Rank (Mean Score)} & \textbf{Mean Score} \\
\midrule
Dutch (nl)           & 1  & 79.96 & 2  & 78.71 \\
Polish (pl)          & 2  & 78.39 & 5  & 77.14 \\
English (en)         & 3  & 77.68 & 3  & 78.08 \\
Portuguese (pt)      & 4  & 77.45 & 6  & 76.20 \\
Italian (it)         & 5  & 75.69 & 1  & 83.22 \\
Lithuanian (lt)      & 6  & 74.10 & 11 & 72.85 \\
Afrikaans (af)       & 7  & 74.02 & 12 & 72.77 \\
Hungarian (hu)       & 8  & 73.45 & 13 & 72.20 \\
French (fr)          & 9  & 70.40 & 7  & 76.15 \\
Indonesian (id)      & 10 & 70.37 & 8  & 75.92 \\
Estonian (et)        & 11 & 70.36 & 4  & 77.89 \\
Bulgarian (bg)       & 12 & 70.16 & 14 & 71.75 \\
Malay (ms)           & 13 & 69.89 & 23 & 62.92 \\
Javanese (jv)        & 14 & 69.66 & 24 & 62.70 \\
Finnish (fi)         & 15 & 69.54 & 22 & 64.96 \\
Spanish (es)         & 16 & 69.42 & 15 & 69.98 \\
Romanian (ro)        & 17 & 69.01 & 17 & 67.76 \\
German (de)          & 18 & 68.35 & 21 & 65.84 \\
Tagalog (tl)         & 19 & 67.66 & 18 & 66.41 \\
Ukrainian (uk)       & 20 & 67.21 & 20 & 65.96 \\
Azerbaijani (az)     & 21 & 66.11 & 28 & 59.15 \\
Vietnamese (vi)      & 22 & 64.96 & 27 & 59.39 \\
Turkish (tr)         & 23 & 63.89 & 19 & 66.20 \\
Swahili (sw)         & 24 & 63.55 & 16 & 68.77 \\
Basque (eu)          & 25 & 61.73 & 9  & 73.46 \\
Russian (ru)         & 26 & 61.34 & 25 & 62.29 \\
Hindi (hi)           & 27 & 60.45 & 33 & 57.04 \\
Greek (el)           & 28 & 60.32 & 29 & 59.14 \\
Arabic (ar)          & 29 & 59.65 & 32 & 57.09 \\
Chinese (zh)         & 30 & 59.61 & 26 & 60.16 \\
Kazakh (kk)          & 31 & 59.52 & 30 & 58.27 \\
Bengali (bn)         & 32 & 59.37 & 43 & 50.94 \\
Marathi (mr)         & 33 & 59.02 & 42 & 51.36 \\
Haitian Creole (ht)  & 34 & 58.45 & 10 & 73.00 \\
Telugu (te)          & 35 & 57.03 & 37 & 55.49 \\
Korean (ko)          & 36 & 55.49 & 35 & 56.27 \\
Persian (fa)         & 37 & 55.19 & 38 & 53.94 \\
Hebrew (he)          & 38 & 55.15 & 40 & 53.90 \\
Georgian (ka)        & 39 & 55.14 & 45 & 48.17 \\
Urdu (ur)            & 40 & 54.16 & 36 & 55.75 \\
Gujarati (gu)        & 41 & 53.26 & 46 & 42.90 \\
Assamese (as)        & 42 & 53.19 & 49 & 41.86 \\
Kannada (kn)         & 43 & 53.01 & 50 & 41.68 \\
Tamil (ta)           & 44 & 52.67 & 41 & 51.95 \\
Punjabi (pa)         & 45 & 52.65 & 47 & 42.29 \\
Malayalam (ml)       & 46 & 51.00 & 51 & 40.64 \\
Thai (th)            & 47 & 48.27 & 44 & 50.58 \\
Quechua (qu)         & 48 & 48.11 & 34 & 56.51 \\
Japanese (ja)        & 49 & 46.88 & 39 & 53.91 \\
Oriya (or)           & 50 & 46.28 & 52 & 34.95 \\
Burmese (my)         & 51 & 43.44 & 31 & 57.59 \\
Yoruba (yo)          & 52 & 43.22 & 48 & 41.98 \\
Wolof (wo)           & 53 & 22.91 & 53 & 27.38 \\
\bottomrule
\end{tabular}
}
\caption{Full ranking of languages by estimated language potential and by mean reported scores, averaged across all models and tasks. The ISO 639-1 code for each language is also provided for reference.}
\label{tab:full_langpot_meanscore}
\end{table}

\section{Model Diagnostics and Robustness Checks}
\label{sec:diagnostics_appendix}

The application of the linear mixed-effects model in our framework relies on several key assumptions, including the normality and homoscedasticity of residuals, as well as the normality of random effects. In this section, we conduct a series of standard model diagnostics to assess the extent to which these assumptions are met in our data. Specifically, we examine the distribution of residuals using Q-Q plots and residual-versus-fitted value plots, and diagnose the distribution of random effects with Q-Q plots. We further conduct formal statistical tests, including the Shapiro-Wilk test~\cite{shapiro1965analysis} for normality and the Levene test~\cite{levene1960robust} for homogeneity of variance across languages.

\begin{figure}[!t]
    \centering
    \includegraphics[width=0.8\linewidth]{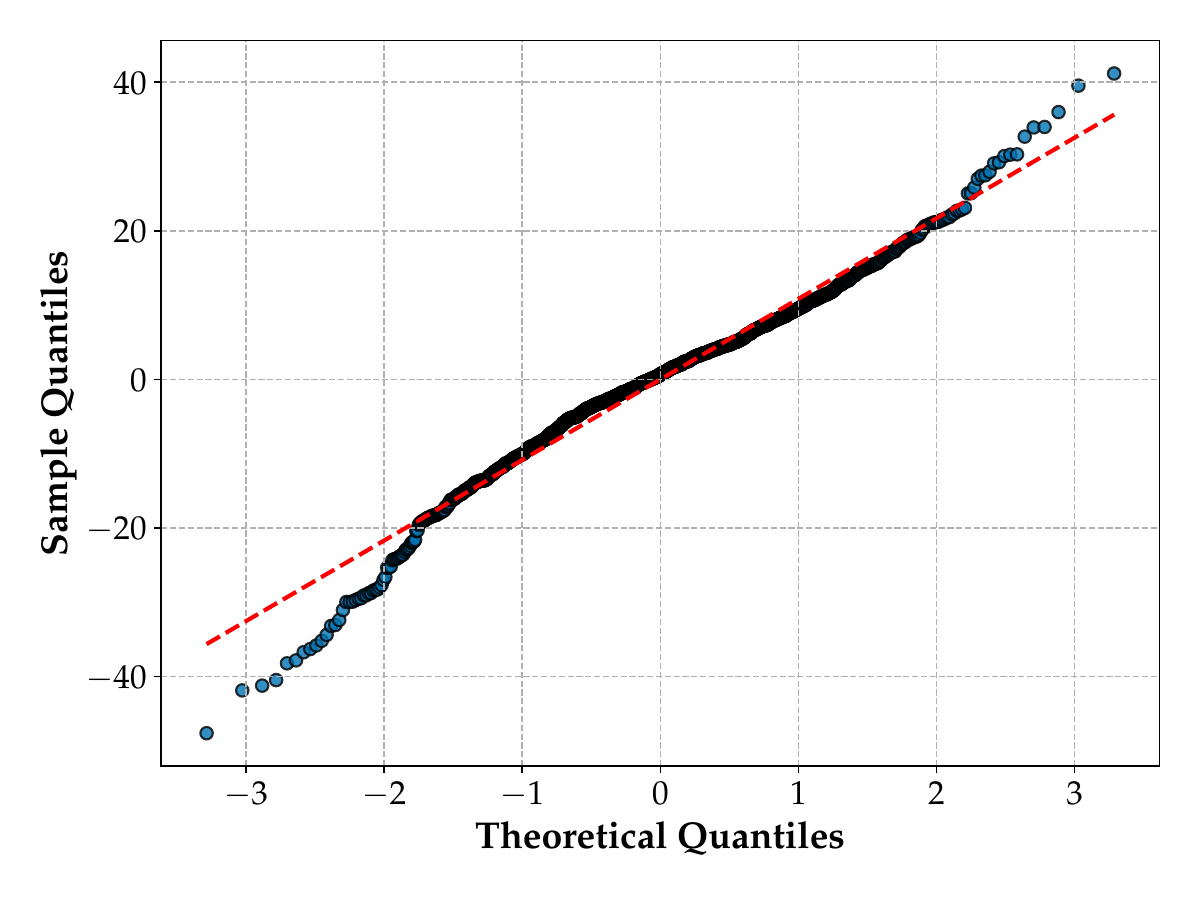}
    \caption{
    Q-Q plot of the residuals from the linear mixed-effects model. Closer alignment of the sample quantiles (dots) to the reference line indicates better adherence to the normality assumption.
    }
    \label{fig:qq_plot_residual_appendix}
\end{figure}

\begin{figure}[!t]
    \centering
    \includegraphics[width=0.8\linewidth]{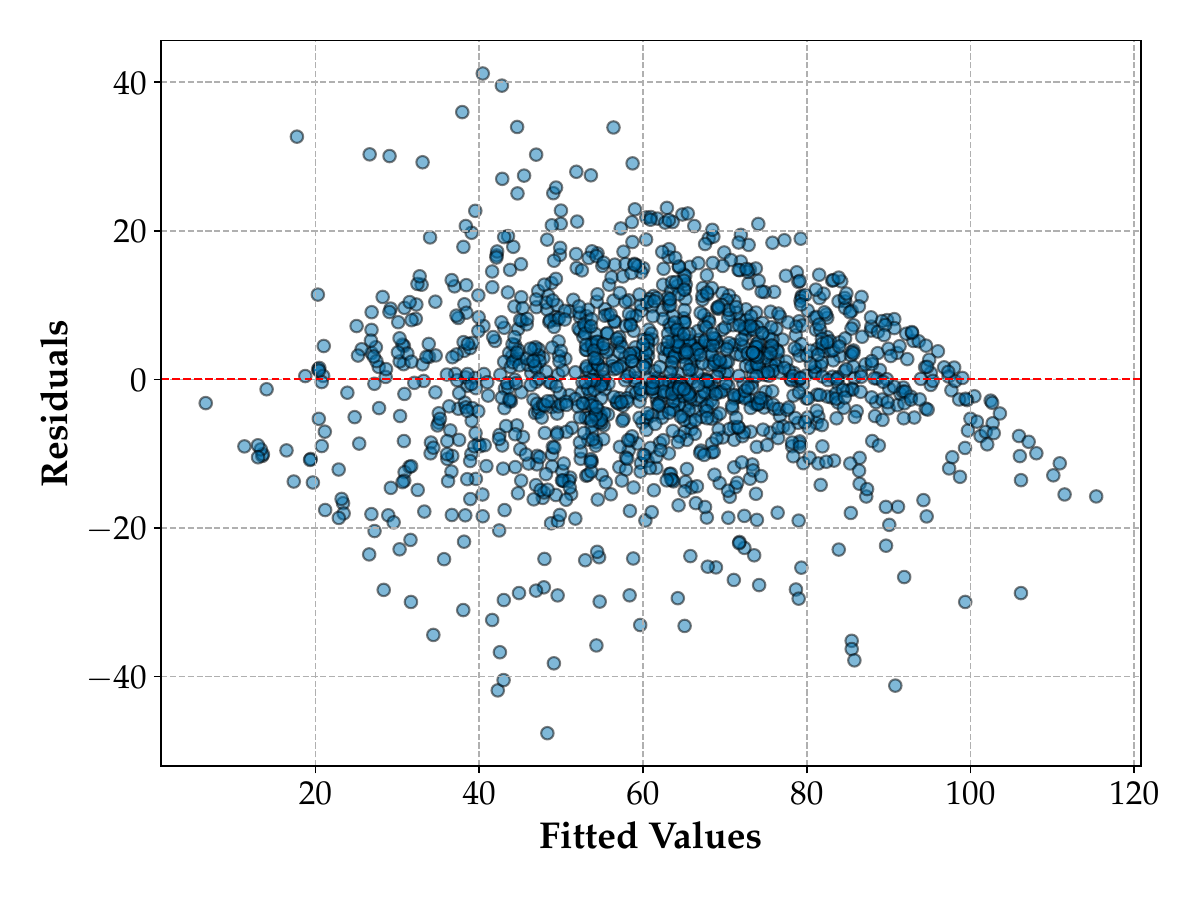}
    \caption{
    Residuals versus fitted values from the linear mixed-effects model. The absence of a clear pattern and constant spread around zero suggest homoscedasticity (constant variance) of the residuals.
    }
    \label{fig:residuals_vs_fitted_appendix}
\end{figure}

\begin{figure}[!t]
    \centering
    \includegraphics[width=0.8\linewidth]{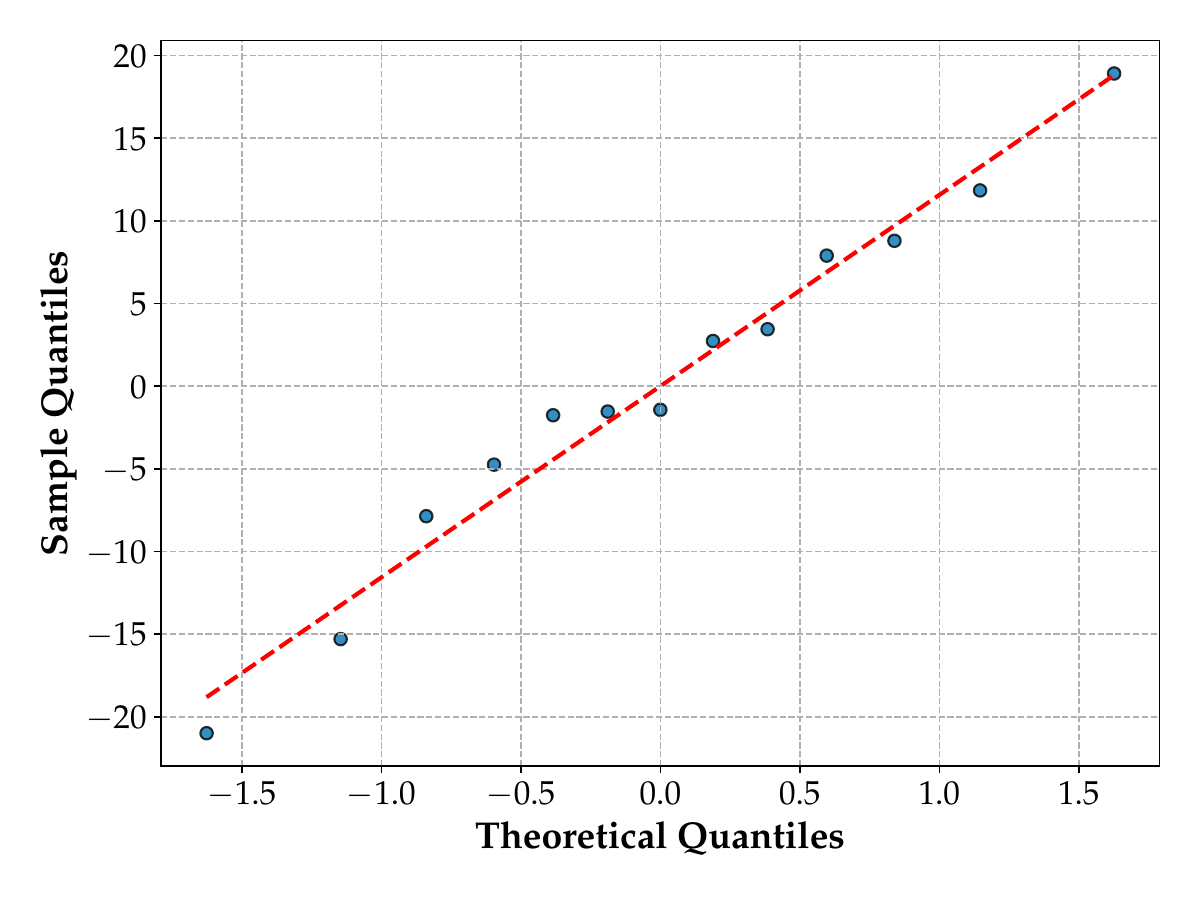}
    \caption{
    Q-Q plot of the random effects from the linear mixed-effects model. Closer alignment of the sample quantiles (dots) to the reference line indicates better adherence to the normality assumption.
    }
    \label{fig:qq_plot_random_appendix}
\end{figure}

\vspace{0.5mm}
\noindent
\textbf{Figure~\ref{fig:qq_plot_residual_appendix}} shows the Q-Q plot of the residuals. Most residuals align well with the reference line, but points at the lower end of the distribution fall below the line, while points at the upper end lie above it. This pattern indicates that the residuals are positively skewed, suggesting that there are more large positive residuals than would be expected under the normality assumption. The Shapiro-Wilk test ($W=0.98$, $p<0.001$) confirms a significant deviation from normality.

\vspace{0.5mm}
\noindent
\textbf{Figure~\ref{fig:residuals_vs_fitted_appendix}} shows the residuals versus fitted values plot. The spread of the residuals is relatively constant across the range of fitted values, with no clear systematic pattern. However, several large-magnitude residuals are present, consistent with the deviations observed in the Q-Q plot. While visual inspection suggests that the assumption of homoscedasticity (constant variance) is reasonable, the Levene test across language groups reveals significant variance heterogeneity ($p < 0.001$), indicating that some languages exhibit more variability in residuals than others. This may reflect language-specific characteristics, as well as potential interaction effects between languages and models, or between languages and tasks, in multilingual evaluation.

\vspace{0.5mm}
\noindent
\textbf{Figure~\ref{fig:qq_plot_random_appendix}} shows the Q-Q plot of the estimated random effects. The alignment of the points with the reference line suggests that the normality assumption for random effects is satisfied. This is further confirmed by the Shapiro-Wilk test ($p = 0.987$), indicating no significant deviation from normality.

Based on the above results, we find that while most model assumptions are reasonably satisfied based on visual inspection, the residuals exhibit deviations from both normality and homoscedasticity. We hypothesise that these violations are primarily attributable to the simplicity of the models employed (as the linear model does not yet account for interactions between effects), the presence of a small number of outliers, and inherent asymmetries in the experimental setups, such as cases where certain languages have test samples but lack corresponding training data (see Table~\ref{tab:app_udpos}). Such deviations are common in large-scale multilingual benchmarking, particularly when fully parallel evaluation setups are difficult to achieve.

\begin{figure}[!t]
    \centering
    \includegraphics[width=\linewidth]{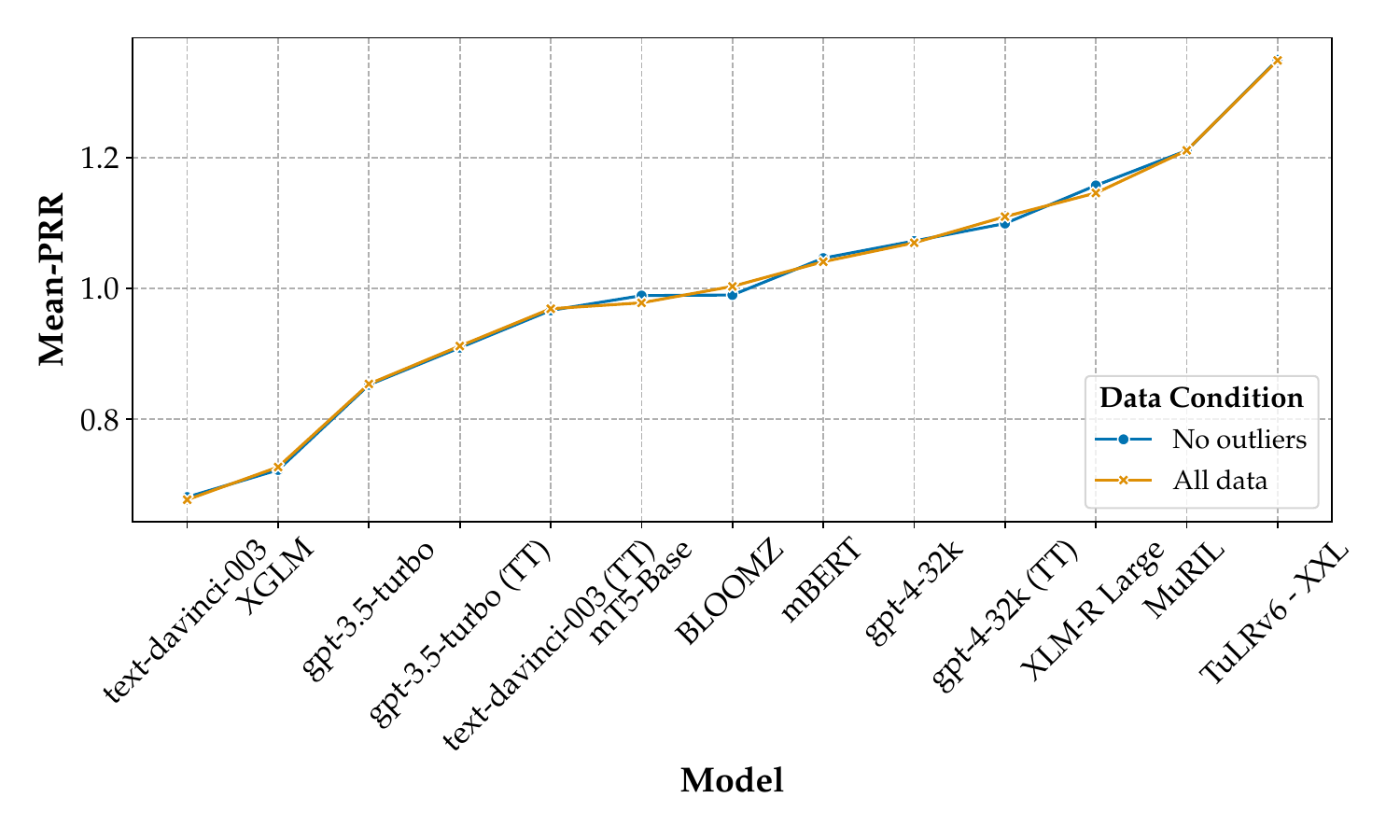}
    \caption{
    Comparison of Mean-PRR (normalised model performance) for each model before and after removing the 10 evaluation records with the largest residuals. The overall model ranking and performance differences remain the same.
    }
    \label{fig:mean_prr_robust_appendix}
\end{figure}

\begin{figure}[!t]
    \centering
    \includegraphics[width=\linewidth]{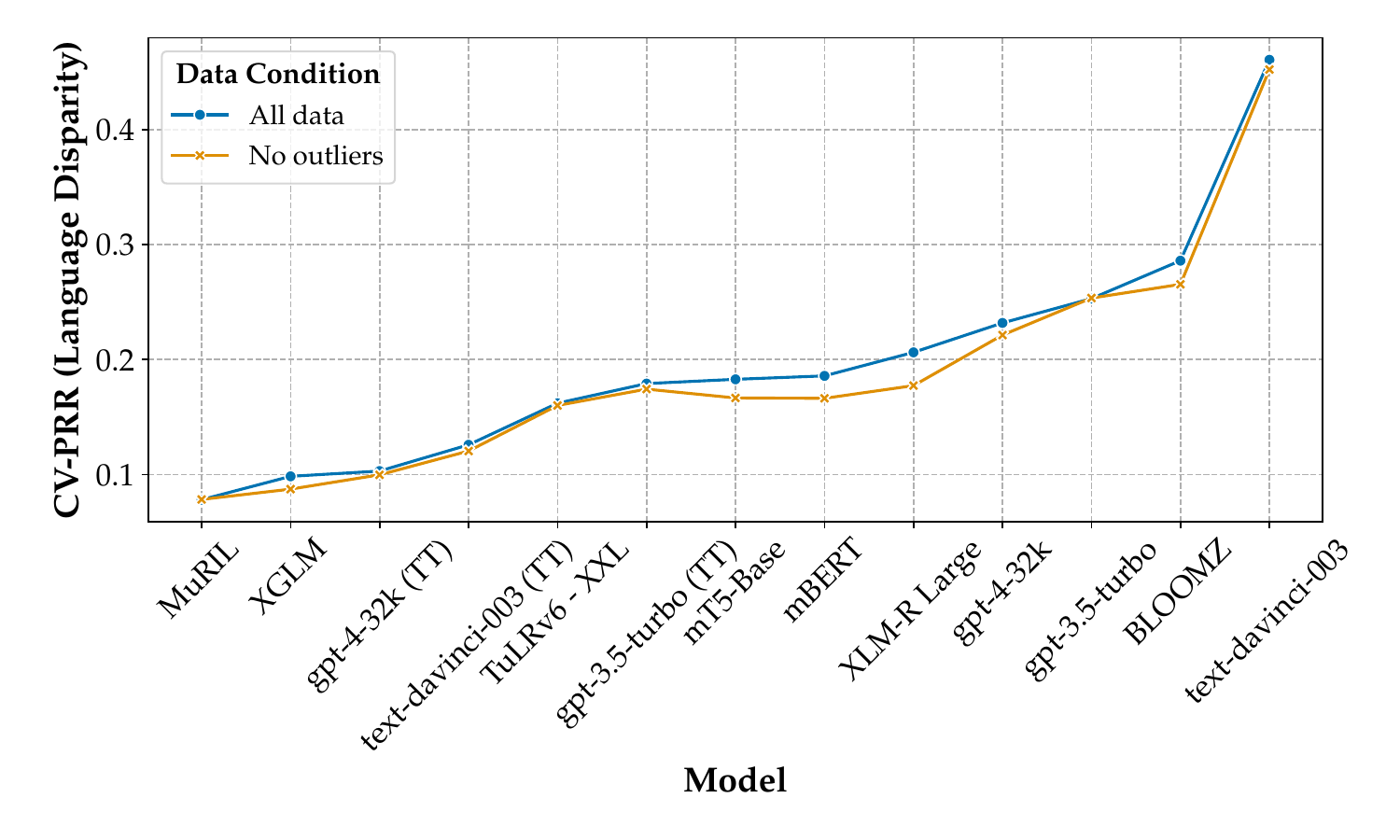}
    \caption{
    Comparison of CV-PRR (model-level language disparity) for each model before and after removing the 10 evaluation records with the largest residuals. The overall ranking and relative disparities remain largely consistent, with the only notable change being a rank swap between \texttt{gpt-3.5-turbo (TT)} and \texttt{mBERT}.
    }
    \label{fig:cv_prr_robust_appendix}
\end{figure}

To evaluate the robustness of our findings to extreme evaluation cases, we repeated the analysis after removing the 10 evaluation records with the largest absolute residuals. This procedure allows us to assess whether our conclusions are overly influenced by a small number of outliers. As shown in \textbf{Figures~\ref{fig:cv_prr_robust_appendix} and~\ref{fig:mean_prr_robust_appendix}}, the results remain largely consistent. The overall model rankings based on Mean-PRR (normalised performance) are unchanged, and only a minor reordering is observed in CV-PRR, where \texttt{gpt-3.5-turbo (TT)} and \texttt{mBERT} swap positions. These findings demonstrate that our framework yields stable and reliable estimates even in the presence of extreme evaluation records.

\vspace{1mm}
\noindent \textbf{Table~\ref{tab:m5_results_appendix}} shows the model-level evaluation results based on our framework for the M5 benchmark~\cite{schneider-sitaram-2024-m5}, which focuses on multilingual and multicultural vision–language tasks.

\begin{table}[]
    \centering
    \begin{tabular}{lcc}
        \toprule
        \textbf{Model} & \textbf{Mean-PRR} & \textbf{CV-PRR} \\
        \midrule
        BakLLaVA       & 0.28 & 4.38 \\
        CogVLM         & 0.27 & 4.39 \\
        GPT-4V         & 0.67 & 3.35 \\
        Gemini Pro V   & 0.52 & 2.77 \\
        InternVL V1.1  & 0.53 & 2.82 \\
        InternVL V1.2+ & 0.63 & 2.43 \\
        LLaVA 1.5 13B  & 0.57 & 2.32 \\
        LLaVA 1.5 7B   & 0.46 & 2.64 \\
        LLaVA 1.6 13B  & 0.61 & 2.59 \\
        LLaVA 1.6 34B  & 0.63 & 2.52 \\
        LLaVA 1.6 7B   & 0.57 & 2.22 \\
        MiniCPM-V      & 0.39 & 3.37 \\
        OmniLMM 12B    & 0.54 & 2.56 \\
        Qwen-VL        & 0.31 & 4.12 \\
        Yi-VL 34B      & 0.36 & 3.60 \\
        Yi-VL 6B       & 0.33 & 3.80 \\
        mBliP BloomZ   & 0.60 & 2.34 \\
        mBliP mT0      & 0.61 & 2.56 \\
        \bottomrule
    \end{tabular}
\caption{Model-level evaluation results based on our framework on the M5 benchmark. Results illustrate that our framework generalises to multilingual vision–language tasks. \textbf{Mean-PRR}: Mean performance realisation ratio; 
    \textbf{CV-PRR}: Coefficient of variation of the PRR.}
    \label{tab:m5_results_appendix}
\end{table}

\section{Original Results from MEGA Benchmark}
\label{sec:mega_result_appendix}

\textbf{Tables~\ref{tab:app_xnli}-\ref{tab:app_xstorycloze}} provide the performance scores used in our case study in \S\ref{sec:case_study}, shown here for completeness. The results are taken from the MEGA benchmark paper of \citet{ahuja-etal-2023-mega}.

\begin{table*}[t]
\centering
\resizebox{\textwidth}{!}{%
\begin{tabular}{lcccccccccccccccc}
\toprule
\textbf{Model} & \textbf{en} & \textbf{ar} & \textbf{bg} & \textbf{de} & \textbf{el} & \textbf{es} & \textbf{fr} & \textbf{hi} & \textbf{ru} & \textbf{sw} & \textbf{th} & \textbf{tr} & \textbf{ur} & \textbf{vi} & \textbf{zh} & \textbf{Avg} \\
\midrule
\multicolumn{16}{l}{\emph{Fine-tuned Baselines}} \\
\midrule
mBERT & 80.8 & 64.3 & 68.0 & 70.0 & 65.3 & 73.5 & 73.4 & 58.9 & 67.8 & 49.7 & 54.1 & 60.9 & 57.2 & 69.3 & 67.8 & 65.4 \\
mT5-Base & 84.7 & 73.3 & 78.6 & 77.4 & 77.1 & 80.3 & 79.1 & 70.8 & 77.1 & 69.4 & 73.2 & 72.8 & 68.3 & 74.2 & 74.1 & 75.4 \\
XLM-R Large & 88.7 & 77.2 & 83.0 & 82.5 & 80.8 & 83.7 & 82.2 & 75.6 & 79.1 & 71.2 & 77.4 & 78.0 & 71.7 & 79.3 & 78.2 & 79.2 \\
TuLRv6 - XXL & 93.3 & 89.0 & 90.6 & 90.0 & 90.2 & 91.1 & 90.7 & 86.2 & 89.2 & 85.5 & 87.5 & 88.4 & 82.7 & 89.0 & 88.4 & 88.8 \\
\midrule
\multicolumn{16}{l}{\emph{Prompt-Based Baselines}} \\
\midrule
BLOOMZ & 67.5 & 60.7 & 46.5 & 54.0 & 47.4 & 61.2 & 61.4 & 56.8 & 53.3 & 50.4 & 43.8 & 42.7 & 50.0 & 61.0 & 56.7 & 54.2 \\
XGLM & 52.6 & 46.4 & 48.9 & 45.6 & 48.7 & 45.8 & 49.4 & 46.8 & 48.6 & 44.5 & 46.6 & 45.4 & 43.4 & 48.5 & 48.8 & 47.3 \\
\midrule
\multicolumn{16}{l}{\emph{OpenAI Models}} \\
\midrule
gpt-3.5-turbo & 76.2 & 59.0 & 63.5 & 67.3 & 65.1 & 70.3 & 67.7 & 55.5 & 62.5 & 56.3 & 54.0 & 62.6 & 49.1 & 60.9 & 62.1 & 62.1 \\
gpt-3.5-turbo (TT) & 76.2 & 62.7 & 67.3 & 69.4 & 67.2 & 69.6 & 69.0 & 59.9 & 63.7 & 55.8 & 59.6 & 63.8 & 54.0 & 63.9 & 62.6 & 64.3 \\
text-davinci-003 & 79.5 & 52.2 & 61.8 & 65.8 & 59.7 & 71.0 & 65.7 & 47.6 & 62.2 & 50.2 & 51.1 & 57.9 & 50.0 & 56.4 & 58.0 & 59.3 \\
text-davinci-003 (TT) & 79.5 & 65.1 & 70.8 & 71.7 & 69.3 & 72.2 & 71.8 & 63.3 & 67.3 & 57.3 & 62.0 & 67.6 & 55.1 & 66.9 & 65.8 & 67.1 \\
gpt-4-32k & 84.9 & 73.1 & 77.3 & 78.8 & 79.0 & 78.8 & 79.5 & 72.0 & 74.3 & 70.9 & 68.8 & 76.3 & 68.1 & 74.3 & 74.6 & 75.4 \\
\bottomrule
\end{tabular}
}
\caption{Detailed results for XNLI by language and model (accuracy). The mapping between the ISO 639-1 code for each language and its language name is provided in Table~\ref{tab:full_langpot_meanscore}.}
\label{tab:app_xnli}
\end{table*}

\begin{table*}[t]
\centering
\resizebox{0.8\textwidth}{!}{%
\begin{tabular}{lcccccccccccccc}
\toprule
\textbf{Model} & \textbf{as} & \textbf{bn} & \textbf{gu} & \textbf{hi} & \textbf{kn} & \textbf{ml} & \textbf{mr} & \textbf{or} & \textbf{pa} & \textbf{ta} & \textbf{te} & \textbf{Avg} \\
\midrule
\multicolumn{13}{l}{\emph{Fine-tuned Baselines}} \\
\midrule
MuRIL & 76.0 & 75.0 & 77.0 & 77.0 & 77.0 & 79.0 & 74.0 & 76.0 & 77.0 & 77.0 & 74.0 & 76.0 \\
\midrule
\multicolumn{13}{l}{\emph{OpenAI Models}} \\
\midrule
gpt-3.5-turbo & 49.5 & 53.6 & 50.6 & 55.5 & 53.9 & 48.4 & 49.9 & 47.4 & 53.6 & 48.2 & 47.4 & 50.7 \\
gpt-3.5-turbo (TT) & 54.3 & 61.6 & 61.8 & 59.6 & 60.8 & 59.9 & 58.7 & 58.5 & 62.3 & 58.3 & 60.8 & 59.7 \\
text-davinci-003 & 48.6 & 52.6 & 51.2 & 56.9 & 49.1 & 48.2 & 49.4 & 46.4 & 50.4 & 45.5 & 47.2 & 49.6 \\
text-davinci-003 (TT) & 56.0 & 66.0 & 64.7 & 62.6 & 63.9 & 61.8 & 60.9 & 60.8 & 64.7 & 61.8 & 63.1 & 62.4 \\
gpt-4-32k & 63.5 & 72.2 & 66.9 & 71.7 & 69.0 & 64.3 & 66.2 & 61.1 & 71.1 & 63.7 & 64.8 & 66.8 \\
\bottomrule
\end{tabular}
}
\caption{Detailed results for IndicXNLI by language and model (accuracy). The mapping between the ISO 639-1 code for each language and its language name is provided in Table~\ref{tab:full_langpot_meanscore}.}
\label{tab:app_indicxnli}
\end{table*}

\begin{table*}[t]
\centering
\resizebox{0.65\textwidth}{!}{%
\begin{tabular}{lcccccccc}
\toprule
\textbf{Model} & \textbf{en} & \textbf{de} & \textbf{es} & \textbf{fr} & \textbf{ja} & \textbf{ko} & \textbf{zh} & \textbf{Avg} \\
\midrule
\multicolumn{9}{l}{\emph{Fine-tuned Baselines}} \\
\midrule
mBERT & 94.0 & 85.7 & 87.4 & 87.0 & 73.0 & 69.6 & 77.0 & 81.9 \\
mT5-Base & 95.4 & 89.4 & 89.6 & 91.2 & 79.8 & 78.5 & 81.1 & 86.4 \\
XLM-R Large & 94.7 & 89.7 & 90.1 & 90.4 & 78.7 & 79.0 & 82.3 & 86.4 \\
TuLRv6 - XXL & 97.2 & 95.1 & 94.8 & 95.6 & 89.4 & 90.4 & 90.4 & 93.2 \\
\midrule
\multicolumn{9}{l}{\emph{Prompt-Based Baselines}} \\
\midrule
BLOOMZ & 89.8 & 84.3 & 88.9 & 87.5 & 74.4 & 85.8 & 65.2 & 82.3\\
\midrule
\multicolumn{9}{l}{\emph{OpenAI Models}} \\
\midrule
gpt-3.5-turbo & 72.4 & 70.6 & 72.0 & 72.1 & 67.2 & 66.5 & 69.2 & 70.0 \\
gpt-3.5-turbo (TT) & 72.4 & 70.8 & 69.7 & 70.1 & 61.9 & 62.5 & 63.1 & 67.2 \\
text-davinci-003 & 72.5 & 70.6 & 72.7 & 70.7 & 60.6 & 61.8 & 60.8 & 67.1 \\
text-davinci-003 (TT) & 72.5 & 69.8 & 70.1 & 71.3 & 65.4 & 65.8 & 65.2 & 68.6 \\
gpt-4-32k & 76.2 & 74.0 & 74.1 & 72.6 & 71.5 & 69.9 & 72.6 & 73.0 \\
\bottomrule
\end{tabular}
}
\caption{Detailed results for PAWS-X by language and model (accuracy). The mapping between the ISO 639-1 code for each language and its language name is provided in Table~\ref{tab:full_langpot_meanscore}.}
\label{tab:app_pawsx}
\end{table*}

\begin{table*}[t]
\centering
\resizebox{0.8\textwidth}{!}{%
\begin{tabular}{lccccccccccc}
\toprule
\textbf{Model} & \textbf{en} & \textbf{et} & \textbf{ht} & \textbf{id} & \textbf{it} & \textbf{qu} & \textbf{sw} & \textbf{ta} & \textbf{th} & \textbf{tr} & \textbf{Avg} \\
\midrule
\multicolumn{12}{l}{\emph{Fine-tuned Baselines}} \\
\midrule
mT5-Base & -- & 50.3 & 49.9 & 49.2 & 49.6 & 50.5 & 50.4 & 49.2 & 50.7 & 49.5 & 49.9 \\
TuLRv6 - XXL & -- & 77.4 & 78.0 & 92.6 & 96.0 & 61.0 & 69.4 & 85.4 & 87.2 & 92.8 & 74.0 \\
\midrule
\multicolumn{12}{l}{\emph{Prompt-Based Baselines}} \\
\midrule
BLOOMZ & 88.0 & 48.0 & 55.0 & 86.0 & 74.0 & 50.0 & 60.0 & 67.0 & 50.0 & 54.0 & 63.2 \\
XGLM & -- & 65.9 & 58.9 & 68.9 & 69.2 & 47.1 & 62.9 & 56.3 & 62.0 & 58.5 & 61.1 \\
\midrule
\multicolumn{12}{l}{\emph{OpenAI Models}} \\
\midrule
gpt-3.5-turbo & 97.8 & 90.6 & 72.0 & 90.4 & 95.2 & 54.6 & 82.0 & 59.0 & 77.6 & 91.0 & 81.0 \\
gpt-3.5-turbo (TT) & 97.8 & 88.2 & 79.4 & 90.8 & 94.4 & 50.0 & 77.6 & 87.0 & 82.2 & 87.8 & 83.5 \\
text-davinci-003 & 98.2 & 87.8 & 75.0 & 91.4 & 96.0 & 54.8 & 63.6 & 53.8 & 66.6 & 87.8 & 77.5 \\
text-davinci-003 (TT) & 98.2 & 89.6 & 82.8 & 93.0 & 94.6 & 50.0 & 82.8 & 87.0 & 84.8 & 89.8 & 85.3 \\
gpt-4-32k & 99.6 & 98.8 & 93.2 & 97.6 & 99.8 & 58.6 & 94.4 & 79.6 & 87.8 & 97.4 & 90.7 \\
gpt-4-32k (TT) & 99.6 & 94.4 & 85.8 & 96.0 & 98.2 & 85.8 & 83.4 & 91.4 & 87.8 & 92.2 & 90.6 \\
\bottomrule
\end{tabular}
}
\caption{Detailed results for XCOPA by language and model (accuracy). The mapping between the ISO 639-1 code for each language and its language name is provided in Table~\ref{tab:full_langpot_meanscore}.}
\label{tab:app_xcopa}
\end{table*}

\begin{table*}[t]
\centering
\resizebox{\textwidth}{!}{%
\begin{tabular}{lcccccccccccc}
\toprule
\textbf{Model} & \textbf{en} & \textbf{ar} & \textbf{de} & \textbf{el} & \textbf{es} & \textbf{hi} & \textbf{ru} & \textbf{th} & \textbf{tr} & \textbf{vi} & \textbf{zh} & \textbf{Avg} \\
\midrule
\multicolumn{13}{l}{\emph{Fine-tuned Baselines}} \\
\midrule
mBERT & 83.5/72.2 & 61.5/45.1 & 70.6/54.0 & 62.6/44.9 & 75.5/56.9 & 59.2/46.0 & 71.3/53.3 & 42.7/33.5 & 55.4/40.1 & 69.5/49.6 & 58.0/48.3 & 64.5/49.4 \\
mT5-Base & 84.6/71.7 & 63.8/44.3 & 73.8/54.5 & 59.6/35.6 & 74.8/56.1 & 60.3/43.4 & 57.8/34.7 & 57.6/45.7 & 67.9/48.2 & 70.7/50.3 & 66.1/54.1 & 67.0/49.0 \\
XLM-R Large & 86.5/75.7 & 68.6/49.0 & 80.4/63.4 & 79.8/61.7 & 82.0/63.9 & 76.7/59.7 & 80.1/64.3 & 74.2/62.8 & 75.9/59.3 & 79.1/59.0 & 59.3/50.0 & 76.6/60.8 \\
TuLRv6 - XXL & 90.1/80.6 & 85.4/69.6 & 86.1/70.4 & 86.3/70.4 & 87.6/71.0 & 85.9/70.5 & 86.8/73.2 & 87.0/81.1 & 84.3/71.0 & 87.6/71.3 & 79.2/73.2 & 86.0/72.9 \\
\midrule
\multicolumn{13}{l}{\emph{Prompt-Based Baselines}} \\
\midrule
BLOOMZ & 92.1/83.8 & 82.8/69.7 & 76.3/60.4 & 49.7/37.6 & 86.8/71.4 & 83.4/72.9 & 65.7/47.2 & 20.5/15.5 & 51.4/37.2 & 86.9/72.7 & 82.4/78.6 & 70.7/58.8 \\
\midrule
\multicolumn{13}{l}{\emph{OpenAI Models}} \\
\midrule
gpt-3.5-turbo & 79.3/58.7 & 59.6/35.1 & 70.6/46.6 & 49.0/22.8 & 70.3/40.8 & 54.0/29.0 & 58.0/31.3 & 41.9/30.4 & 61.8/35.0 & 69.1/42.4 & 50.4/48.3 & 60.4/38.2 \\
text-davinci-003 & 77.2/61.8 & 36.8/22.5 & 55.2/39.7 & 31.8/19.7 & 61.8/41.3 & 19.9/10.0 & 29.4/17.6 & 11.5/8.7 & 44.8/29.2 & 41.7/25.4 & 35.6/32.8 & 40.5/28.1 \\
gpt-4-32k & 83.2/65.6 & 67.8/42.4 & 71.9/48.7 & 62.3/36.6 & 77.5/50.7 & 63.9/36.7 & 63.8/35.8 & 54.6/42.0 & 70.8/46.6 & 75.8/49.7 & 60.0/57.5 & 68.3/46.6 \\
\bottomrule
\end{tabular}
}
\caption{Detailed results for XQuAD by language and model (F1 Score / Exact Match). The mapping between the ISO 639-1 code for each language and its language name is provided in Table~\ref{tab:full_langpot_meanscore}.}
\label{tab:app_xquad}
\end{table*}

\begin{table*}[t]
\centering
\resizebox{\textwidth}{!}{%
\begin{tabular}{lcccccccccc}
\toprule
\textbf{Model} & \textbf{en} & \textbf{ar} & \textbf{bn} & \textbf{fi} & \textbf{id} & \textbf{ko} & \textbf{ru} & \textbf{sw} & \textbf{te} & \textbf{Avg} \\
\midrule
\multicolumn{11}{l}{\emph{Fine-tuned Baselines}} \\
\midrule
mBERT         & 75.3/63.6 & 62.2/42.8 & 49.3/32.7 & 59.7/45.3 & 64.8/45.8 & 58.8/50.0 & 60.0/38.8 & 57.5/37.9 & 49.6/38.4 & 59.7/43.9 \\
mT5-Base      & 71.8/60.9 & 67.1/50.4 & 40.7/22.1 & 67.0/52.2 & 71.3/54.5 & 49.5/37.7 & 54.9/32.6 & 60.4/43.9 & 40.6/31.1 & 58.1/42.8 \\
XLM-R Large   & 71.5/56.8 & 67.6/40.4 & 64.0/47.8 & 70.5/53.2 & 77.4/61.9 & 31.9/10.9 & 67.0/42.1 & 66.1/48.1 & 70.1/43.6 & 65.1/45.0 \\
TuLRv6 - XXL  & 85.4/76.4 & 84.1/70.4 & 86.9/79.6 & 83.8/72.8 & 88.8/77.9 & 78.5/67.8 & 81.9/68.6 & 87.2/79.6 & 85.2/71.6 & 84.6/73.8 \\
\midrule
\multicolumn{11}{l}{\emph{Prompt-Based Baselines}} \\
\midrule
BLOOMZ        & 82.4/70.9 & 81.9/62.2 & 87.8/82.3 & 43.6/28.6 & 85.0/71.0 & 52.3/43.1 & 67.4/51.5 & 86.0/77.2 & 90.3/81.6 & 75.2/63.2 \\
\midrule
\multicolumn{11}{l}{\emph{OpenAI Models}} \\
\midrule
gpt-3.5-turbo    & 54.8/30.7 & 50.9/24.2 & 60.7/32.7 & 66.6/49.0 & 67.2/43.4 & 59.7/45.3 & 45.8/20.0 & 64.3/47.7 & 70.9/53.1 & 60.1/38.4 \\
text-davinci-003 & 73.7/59.1 & 56.2/38.7 & 16.1/10.6 & 70.3/58.8 & 68.6/51.2 & 40.6/32.2 & 42.3/28.9 & 74.1/62.3 & 5.8/3.0   & 49.8/38.3 \\
gpt-4-32k        & 72.9/51.4 & 60.8/32.7 & 68.0/42.5 & 75.4/57.7 & 80.8/61.1 & 69.7/58.5 & 61.4/30.5 & 81.8/68.7 & 72.5/54.9 & 71.5/50.9 \\
\bottomrule
\end{tabular}
}
\caption{Detailed results for TyDiQA-GoldP by language and model (F1 Score / Exact Match). The mapping between the ISO 639-1 code for each language and its language name is provided in Table~\ref{tab:full_langpot_meanscore}.}
\label{tab:app_tydiqa}
\end{table*}

\begin{table*}[t]
\centering
\resizebox{0.9\textwidth}{!}{%
\begin{tabular}{lcccccccc}
\toprule
\textbf{Model} & \textbf{en} & \textbf{ar} & \textbf{de} & \textbf{es} & \textbf{hi} & \textbf{vi} & \textbf{zh} & \textbf{Avg} \\
\midrule
\multicolumn{9}{l}{\emph{Fine-tuned Baselines}} \\
\midrule
mBERT         & 80.2/67.0 & 52.3/34.6 & 59.0/43.8 & 67.4/49.2 & 50.2/35.3 & 61.2/40.7 & 59.6/38.6 & 61.4/44.2 \\
mT5-Base      & 81.7/66.9 & 57.1/36.9 & 62.1/43.2 & 67.1/47.2 & 55.4/37.9 & 65.9/44.1 & 61.6/38.6 & 64.4/45.0 \\
XLM-R Large   & 83.5/70.6 & 66.6/47.1 & 70.1/54.9 & 74.1/56.6 & 70.6/53.1 & 74.0/52.9 & 62.1/37.0 & 71.6/53.2 \\
TuLRv6 - XXL  & 86.6/74.4 & 76.2/56.5 & 80.2/67.0 & 81.7/65.1 & 82.2/64.8 & 82.3/63.2 & 78.1/56.5 & 81.0/63.9 \\
\midrule
\multicolumn{9}{l}{\emph{OpenAI Models}} \\
\midrule
gpt-3.5-turbo        & 72.8/53.2 & 48.5/23.9 & 51.0/29.6 & 53.8/29.4 & 50.7/28.9 & 58.9/35.1 & 56.7/29.4 & 56.1/32.8 \\
gpt-3.5-turbo (TT)   & 72.8/53.2 & 37.8/18.4 & 44.3/26.2 & 54.1/31.8 & 37.3/20.0 & 41.6/22.5 & 36.5/17.2 & 46.4/27.0 \\
text-davinci-003     & 74.8/59.0 & 38.4/21.7 & 57.7/38.1 & 62.9/37.8 & 24.9/14.1 & 47.7/29.7 & 32.3/31.7 & 48.4/33.1 \\
text-davinci-003 (TT)& 74.8/59.0 & 48.2/25.6 & 53.5/33.9 & 62.9/40.9 & 49.2/28.7 & 51.0/30.4 & 45.2/24.1 & 55.0/34.7 \\
gpt-4-32k            & 80.3/62.8 & 59.1/33.5 & 64.7/44.4 & 70.0/45.9 & 57.3/35.6 & 72.2/49.0 & 67.1/38.4 & 67.2/44.2 \\
\bottomrule
\end{tabular}
}
\caption{Detailed results for MLQA by language and model (F1 Score / Exact Match). The mapping between the ISO 639-1 code for each language and its language name is provided in Table~\ref{tab:full_langpot_meanscore}.}
\label{tab:app_mlqa}
\end{table*}

\begin{table*}[t]
\centering
\resizebox{\textwidth}{!}{%
\begin{tabular}{lccccccccccccc}
\toprule
\textbf{Model} & \textbf{as} & \textbf{bn} & \textbf{gu} & \textbf{hi} & \textbf{kn} & \textbf{ml} & \textbf{mr} & \textbf{or} & \textbf{pa} & \textbf{ta} & \textbf{te} & \textbf{Avg} \\
\midrule
\multicolumn{13}{l}{\emph{Fine-tuned Baselines}} \\
\midrule
BLOOMZ & 40.6/31.7 & 42.9/36.6 & 37.2/29.9 & 44.0/45.1 & 37.8/26.6 & 30.5/28.4 & 39.2/33.0 & 25.4/22.0 & 26.4/33.5 & 39.7/35.9 & 38.9/34.7 & 36.6/32.5 \\
\midrule
\multicolumn{13}{l}{\emph{OpenAI Models}} \\
\midrule
gpt-3.5-turbo      & 35.3/21.4 & 49.5/30.2 & 40.5/25.5 & 55.9/39.3 & 35.3/20.4 & 30.0/19.2 & 50.0/32.0 & 22.1/12.7 & 35.8/15.1 & 32.7/21.6 & 32.9/19.7 & 38.2/23.4 \\
text-davinci-003   & 6.7/3.2 & 10.3/5.8 & 5.4/3.5 & 16.8/11.8 & 7.1/3.9 & 3.6/2.3 & 14.6/8.5 & 6.9/3.4 & 10.7/4.1 & 4.2/2.5 & 6.8/3.6 & 8.4/4.8 \\
gpt-4-32k          & 58.8/40.4 & 67.1/47.4 & 59.4/42.4 & 75.2/62.2 & 47.1/31.6 & 48.3/33.7 & 60.7/43.1 & 29.9/16.7 & 56.1/34.1 & 54.0/39.7 & 47.9/27.8 & 55.0/38.1 \\
\bottomrule
\end{tabular}
}
\caption{Detailed results for IndicQA by language and model (F1 Score / Exact Match). The mapping between the ISO 639-1 code for each language and its language name is provided in Table~\ref{tab:full_langpot_meanscore}.}
\label{tab:app_indicqa}
\end{table*}

\begin{table*}[t]
\centering
\resizebox{\textwidth}{!}{%
\begin{tabular}{lcccccccccccccccccccc}
\toprule
\textbf{Model} & \textbf{en} & \textbf{af} & \textbf{ar} & \textbf{bg} & \textbf{de} & \textbf{el} & \textbf{es} & \textbf{et} & \textbf{eu} & \textbf{fa} & \textbf{fi} & \textbf{fr} & \textbf{he} & \textbf{hi} & \textbf{hu} & \textbf{id} & \textbf{it} & \textbf{ja} & \textbf{kk} \\
\midrule
\multicolumn{20}{l}{\emph{Fine-tuned Baselines}} \\
\midrule
mBERT & 96.4 & 86.7 & 50.0 & 84.7 & 88.7 & 80.9 & 86.6 & 79.9 & 62.1 & 65.5 & 73.3 & 81.2 & 55.5 & 66.0 & 78.6 & 74.2 & 87.8 & 47.2 & 70.4 \\
XLM-R Large & 97.0 & 89.2 & 63.0 & 88.3 & 91.2 & 86.5 & 89.2 & 87.3 & 74.9 & 70.8 & 82.7 & 86.7 & 67.5 & 75.2 & 83.4 & 75.7 & 89.2 & 29.3 & 78.3 \\
\midrule
\multicolumn{20}{l}{\emph{OpenAI Models}} \\
\midrule
gpt-3.5-turbo & 78.5 & 74.3 & 38.3 & 79.1 & 80.7 & 47.1 & 34.8 & 76.0 & 72.0 & 46.7 & 79.5 & 78.0 & 53.8 & 50.7 & 65.4 & 63.6 & 75.4 & 47.4 & 64.8 \\
gpt-4-32k & 84.1 & 77.6 & 42.0 & 83.1 & 86.3 & 49.8 & 68.4 & 80.2 & 79.3 & 46.4 & 82.7 & 85.4 & 60.4 & 52.2 & 68.3 & 68.6 & 84.1 & 60.2 & 71.8 \\
\midrule
& \textbf{ko} & \textbf{lt} & \textbf{mr} & \textbf{nl} & \textbf{pl} & \textbf{pt} & \textbf{ro} & \textbf{ru} & \textbf{ta} & \textbf{te} & \textbf{th} & \textbf{tl} & \textbf{tr} & \textbf{uk} & \textbf{ur} & \textbf{vi} & \textbf{wo} & \textbf{yo} & \textbf{zh} & \textbf{avg} \\
\midrule
\multicolumn{20}{l}{\emph{Fine-tuned Baselines}} \\
\midrule
mBERT & 51.7 & 78.8 & 68.7 & 88.6 & 80.7 & 88.0 & 71.5 & 82.4 & 58.5 & 75.2 & 41.3 & 80.5 & 70.5 & 80.6 & 56.6 & 55.4 & 0.0 & 56.6 & 59.6 & 71.9 \\
XLM-R Large & 57.1 & 84.2 & 81.8 & 89.5 & 86.8 & 90.2 & 82.6 & 87.3 & 64.0 & 84.2 & 48.5 & 92.4 & 81.2 & 85.8 & 70.8 & 58.5 & 0.0 & 24.8 & 44.1 & 76.2 \\
\midrule
\multicolumn{20}{l}{\emph{OpenAI Models}} \\
\midrule
gpt-3.5-turbo & 39.0 & 71.3 & 57.9 & 78.3 & 81.7 & 76.7 & 66.7 & 69.9 & 32.6 & 79.8 & 25.5 & 54.3 & 77.2 & 58.9 & 39.9 & 57.7 & 50.4 & 7.0 & 57.2 & 60.2 \\
gpt-4-32k & 51.2 & 73.7 & 79.1 & 81.8$^{\dagger}$ & 80.7 & 81.0 & 66.3$^{\dagger}$ & 74.7 & 34.7 & 84.6 & 31.2$^{\dagger}$ & 58.4$^{\dagger}$ & 77.0 & 61.9 & 41.3 & 64.7 & 59.1 & 33.8$^{\dagger}$ & 63.5 & 66.6 \\
\bottomrule
\end{tabular}
}
\caption{Comparison of model F1 scores on UDPOS for each language. All numbers are monolingual results except the ones marked with $\dagger$, which indicate zero-shot cross-lingual results (due to absence of training data in those languages). The mapping between the ISO 639-1 code for each language and its language name is provided in Table~\ref{tab:full_langpot_meanscore}.}
\label{tab:app_udpos}
\end{table*}

\begin{table*}[t]
\centering
\resizebox{\textwidth}{!}{%
\begin{tabular}{lcccccccccccccccccccccccccc}
\toprule
\textbf{Model} & \textbf{en} & \textbf{af} & \textbf{ar} & \textbf{az} & \textbf{bg} & \textbf{bn} & \textbf{de} & \textbf{el} & \textbf{es} & \textbf{et} & \textbf{eu} & \textbf{fa} & \textbf{fi} & \textbf{fr} & \textbf{gu} & \textbf{he} & \textbf{hi} & \textbf{hu} & \textbf{id} & \textbf{it} & \textbf{ja} & \textbf{jv} & \textbf{ka} & \textbf{kk} \\
\midrule
\multicolumn{26}{l}{\emph{Fine-tuned Baselines}} \\
\midrule
mBERT & 86.4 & 76.1 & 42.9 & 65.5 & 76.7 & 69.7 & 79.5 & 70.9 & 75.3 & 75.8 & 64.4 & 40.0 & 76.6 & 79.6 & 51.3 & 56.2 & 65.9 & 76.1 & 61.0 & 81.3 & 29.2 & 62.4 & 65.1 & 50.3 \\
XLM-R Large & 85.4 & 78.6 & 47.3 & 69.4 & 80.9 & 74.7 & 80.7 & 79.2 & 71.8 & 78.7 & 61.6 & 55.2 & 79.6 & 79.8 & 62.7 & 55.5 & 70.9 & 80.2 & 51.8 & 80.3 & 18.5 & 61.9 & 70.9 & 54.4 \\
\midrule
\multicolumn{26}{l}{\emph{OpenAI Models}} \\
\midrule
gpt-3.5-turbo & 43.2 & 43.8 & 45.4 & 42.1 & 51.6 & 40.3 & 52.7 & 41.0 & 60.2 & 58.7 & 31.5 & 39.3 & 59.1 & 50.7 & 18.4 & 34.3 & 45.5 & 53.7 & 58.4 & 60.0 & 7.4 & 57.7 & 25.1 & 30.9 \\
gpt-4-32k & 49.7 & 55.9 & 59.4 & 59.6 & 62.6 & 52.7 & 69.2 & 54.4 & 68.6 & 74.4 & 57.8 & 67.6 & 71.1 & 68.5 & 23.8 & 48.0 & 59.4 & 71.9 & 72.7 & 72.8 & 9.2 & 68.8 & 31.6 & 45.3 \\
\midrule
& \textbf{ko} & \textbf{lt} & \textbf{ml} & \textbf{mr} & \textbf{ms} & \textbf{my} & \textbf{nl} & \textbf{pa} & \textbf{pl} & \textbf{pt} & \textbf{qu} & \textbf{ro} & \textbf{ru} & \textbf{sw} & \textbf{ta} & \textbf{te} & \textbf{th} & \textbf{tl} & \textbf{tr} & \textbf{uk} & \textbf{ur} & \textbf{vi} & \textbf{yo} & \textbf{zh} & \textbf{avg} \\
\midrule
\multicolumn{26}{l}{\emph{Fine-tuned Baselines}} \\
\midrule
mBERT & 59.5 & 75.8 & 53.0 & 57.0 & 67.1 & 45.7 & 81.0 & 30.5 & 79.2 & 80.4 & 58.5 & 74.0 & 63.9 & 71.4 & 50.7 & 48.9 & 0.4 & 72.6 & 73.4 & 69.7 & 35.4 & 74.5 & 45.8 & 42.5 & 62.3 \\
XLM-R Large & 59.2 & 75.8 & 60.2 & 63.4 & 68.5 & 55.2 & 83.2 & 49.4 & 79.3 & 79.9 & 58.5 & 78.7 & 71.9 & 68.9 & 58.4 & 53.8 & 0.7 & 74.7 & 80.3 & 78.0 & 60.3 & 78.3 & 37.0 & 26.6 & 65.2 \\
\midrule
\multicolumn{26}{l}{\emph{OpenAI Models}} \\
\midrule
gpt-3.5-turbo & 27.9 & 51.9 & 25.2 & 34.4 & 52.0 & 8.7 & 59.4 & 36.7 & 58.4 & 48.9 & 41.9 & 42.7 & 29.4 & 57.7 & 26.0 & 22.0 & 1.7 & 36.5 & 50.5 & 34.4 & 35.7 & 33.5 & 56.9 & 13.3 & 40.3 \\
gpt-4-32k & 51.4 & 71.3 & 35.6 & 47.4 & 64.1 & 16.3 & 67.9 & 49.8 & 70.3 & 64.5 & 69.8 & 59.6 & 64.8 & 68.9 & 36.9 & 33.0 & 2.5 & 61.9 & 72.9 & 58.4 & 69.6 & 58.4 & 73.9 & 18.5 & 55.5 \\
\bottomrule
\end{tabular}
}
\caption{Comparison of model F1 scores on PAN-X (NER) for each language. The mapping between the ISO 639-1 code for each language and its language name is provided in Table~\ref{tab:full_langpot_meanscore}.}
\label{tab:app_panx}
\end{table*}

\begin{table*}[t]
\centering
\resizebox{0.8\textwidth}{!}{%
\begin{tabular}{lcccccccccccc}
\toprule
\textbf{Model} & \textbf{ar} & \textbf{en} & \textbf{es} & \textbf{eu} & \textbf{hi} & \textbf{id} & \textbf{my} & \textbf{ru} & \textbf{sw} & \textbf{te} & \textbf{zh} & \textbf{avg} \\
\midrule
\multicolumn{12}{l}{\emph{Prompt-Based Baselines}} \\
\midrule
BLOOMZ & 79.7 & 95.7 & 87.3 & 70.5 & 79.9 & 85.6 & 49.9 & 67.3 & 65.3 & 67.4 & 90.0 & 76.2 \\
XGLM & 59.8 & 75.9 & 69.2 & 63.8 & 62.5 & 70.8 & 61.2 & 72.4 & 65.2 & 63.4 & 67.7 & 66.5 \\
\midrule
\multicolumn{12}{l}{\emph{OpenAI Models}} \\
\midrule
gpt-3.5-turbo & 92.5 & 96.8 & 95.8 & 78.4 & 91.1 & 95.0 & 57.2 & 96.6 & 92.3 & 73.1 & 95.6 & 87.7 \\
gpt-3.5-turbo (TT) & 94.3 & 96.8 & 96.1 & 92.5 & 94.7 & 95.2 & 88.6 & 96.2 & 88.7 & 93.6 & 95.6 & 93.9 \\
text-davinci-003 & 87.4 & 98.3 & 97.6 & 78.1 & 77.8 & 96.4 & 47.4 & 94.2 & 78.1 & 57.6 & 95.0 & 82.5 \\
text-davinci-003 (TT) & 95.0 & 98.3 & 96.2 & 94.1 & 95.1 & 95.9 & 90.1 & 96.9 & 90.7 & 94.3 & 96.2 & 94.8 \\
gpt-4-32k & 99.1 & 99.6 & 99.5 & 97.6 & 98.8 & 99.0 & 77.6 & 99.1 & 98.4 & 93.4 & 99.2 & 96.5 \\
gpt-4-32k (TT) & 97.7 & 99.6 & 98.7 & 96.8 & 97.9 & 98.1 & 93.2 & 99.2 & 93.6 & 96.4 & 98.3 & 97.0 \\
\bottomrule
\end{tabular}
}
\caption{
Comparison of model accuracy (\%) on XStoryCloze for each language. 
The mapping between the ISO 639-1 code for each language and its language name is provided in Table~\ref{tab:full_langpot_meanscore}.}
\label{tab:app_xstorycloze}
\end{table*}

\end{document}